\documentclass{article}
\usepackage{arxiv}
\usepackage{hyperref}       
\usepackage{url}
\usepackage{booktabs}       
\usepackage{amsfonts}
\usepackage{nicefrac}
\usepackage{microtype}   
\usepackage{lipsum}
\usepackage{graphicx}
\usepackage{verbatim}
\usepackage{multirow}
\usepackage{multicol}
\usepackage{color}
\usepackage{latexsym}
\usepackage{graphicx}
\usepackage{xcolor}
\usepackage{colortbl,booktabs}
\usepackage{tabu}
\usepackage{float}
\usepackage{amsmath}
\usepackage{wrapfig}
\usepackage{multirow}
\usepackage{makecell}
\usepackage{caption}
\usepackage{soul}
\usepackage{CJKutf8}
\usepackage[utf8]{inputenc} % allow utf-8 input
\usepackage[T1]{fontenc} % use 8-bit T1 fonts
\usepackage{CJK}
\makeatletter
\def\UrlAlphabet{%
      \do\a\do\b\do\c\do\d\do\e\do\f\do\g\do\h\do\i\do\j%
      \do\k\do\l\do\m\do\n\do\o\do\p\do\q\do\r\do\s\do\t%
      \do\u\do\v\do\w\do\x\do\y\do\z\do\A\do\B\do\C\do\D%
      \do\E\do\F\do\G\do\H\do\I\do\J\do\K\do\L\do\M\do\N%
      \do\O\do\P\do\Q\do\R\do\S\do\T\do\U\do\V\do\W\do\X%
      \do\Y\do\Z}
\def\UrlDigits{\do\1\do\2\do\3\do\4\do\5\do\6\do\7\do\8\do\9\do\0}
\g@addto@macro{\UrlBreaks}{\UrlOrds}
\g@addto@macro{\UrlBreaks}{\UrlAlphabet}

\g@addto@macro{\UrlBreaks}{\UrlDigits}
\makeatother

\title{\textsc{ERNIE 3.0 Titan: Exploring Larger-scale Knowledge Enhanced Pre-training for Language Understanding and Generation}
}

\author{Shuohuan Wang\thanks{~~Equal contribution.} $^{\dagger}$\quad\quad Yu Sun$^{*\dagger}$\quad\quad Yang Xiang$^{*\mathsection}$\quad\quad Zhihua Wu$^{\dagger}$\quad\quad Siyu Ding$^{\dagger}$
\\[10pt]
{\bf Weibao Gong$^{\dagger}$\quad\quad Shikun Feng$^{\dagger}$\quad\quad Junyuan Shang$^{\dagger}$\quad\quad Yanbin Zhao$^{\dagger}$\quad\quad Chao Pang$^{\dagger}$\quad Jiaxiang Liu$^{\dagger}$} 
\\[10pt]
{\bf Xuyi Chen$^{\dagger}$\quad\quad Yuxiang Lu$^{\dagger}$\quad\quad Weixin Liu$^{\dagger}$\quad\quad Xi Wang$^{\dagger}$\quad\quad Yangfan Bai$^{\dagger}$\quad\quad Qiuliang Chen$^{\dagger}$}
\\[10pt]
{\bf Li Zhao$^{\dagger}$\quad\quad Shiyong Li$^{\dagger}$\quad\quad Peng Sun$^{\dagger}$\quad\quad Dianhai Yu$^{\dagger}$\quad\quad Yanjun Ma$^{\dagger}$\quad\quad Hao Tian$^{\dagger}$}
\\[10pt]
{\bf \quad\quad Hua Wu$^{\dagger}$ \quad\quad Tian Wu$^{\dagger}$ \quad\quad Wei Zeng$^{\mathsection}$\quad\quad Ge Li$^{\mathsection}$ \quad\quad Wen Gao$^{\mathsection}$ \quad\quad Haifeng Wang$^{\dagger}$} 
\\[17pt]
{\bf $^{\dagger}$ Baidu Inc.} \\
{\tt \{wangshuohuan, sunyu02\}@baidu.com}
\\[10pt]
{\bf $^{\mathsection}$ Peng Cheng Laboratory} \\
{\tt \{xiangy\}@pcl.ac.cn} 
}

\begin{document}
\maketitle
\begin{abstract} %
Pre-trained language models have achieved state-of-the-art results in various Natural Language Processing (NLP) tasks.
GPT-3~\cite{gpt-3} has shown that scaling up pre-trained language models can further exploit their enormous potential.
A unified framework named ERNIE 3.0~\cite{sun2021ernie} was recently proposed for pre-training large-scale knowledge enhanced models and trained a model with 10 billion parameters. ERNIE 3.0 outperformed the state-of-the-art models on various NLP tasks. 
In order to explore the performance of scaling up ERNIE 3.0, we train a hundred-billion-parameter model called ERNIE 3.0 Titan with up to 260 billion parameters on the PaddlePaddle~\cite{ma2019paddlepaddle} platform. Furthermore, we design a self-supervised
adversarial loss and a controllable language modeling loss to make ERNIE 3.0 Titan generate credible and controllable texts. To reduce the computation overhead and carbon emission, we propose an online distillation framework for ERNIE 3.0 Titan, where the teacher model will teach students and train itself simultaneously.
ERNIE 3.0 Titan is the largest Chinese dense pre-trained model so far.
Empirical results show that the ERNIE 3.0 Titan outperforms the state-of-the-art models on 68 NLP datasets.

\end{abstract}
\section{Introduction} 
Pre-trained language models such as ELMo~\cite{peters2018deep}, GPT~\cite{radford2018improving}, BERT~\cite{devlin2018bert}, and ERNIE~\cite{sun2019ernie} have proven effective for improving the performances of various natural language understanding and generation tasks. Pre-trained language models are generally learned on a large amount of text data in a self-supervised manner and then fine-tuned on downstream tasks or directly deployed through zero-shot learning without task-specific fine-tuning. 
Such pre-trained language models start to serve as foundation models and bring a new paradigm for natural language processing tasks. This new paradigm changes the focus of NLP research from designing specialized models for different tasks to studying pre-trained language models and using them in various tasks. Recent advances such as GPT-3~\cite{gpt-3} have demonstrated the promising trend towards scaling up pre-trained language models with billions of parameters. These studies show surprising potentials by scaling up pre-trained models.

Most of existing large-scale models were pre-trained on plain texts without integrating knowledge. ERNIE 3.0~\cite{sun2021ernie} tried to incorporate knowledge such as linguistic knowledge and world knowledge into large-scale pre-trained language models. ERNIE 3.0 pre-trained Transformers on massive unstructured texts and knowledge graphs to learn different levels of knowledge, such as lexical, syntactic, and semantic information. ERNIE 3.0 can handle both natural language understanding tasks and natural language generation tasks through zero-shot learning, few-shot learning, or fine-tuning. Furthermore, it supports introducing various customized tasks at any time. These tasks share the same encoding networks that are pre-trained through multi-task learning. This method makes it possible to encode lexical, syntactic, and semantic information across different tasks.

This work explores the performance of knowledge-enhanced pre-trained models with larger-scale parameters based on the ERNIE 3.0 framework. We train a Chinese dense pre-trained language model with 260 billion parameters (named as \textbf{ERNIE 3.0 Titan}) on the PaddlePaddle platform. Although large-scale language models like GPT-3 have shown promising text generation capabilities, it is still challenging for users to control the generation results and obtain generated texts that are factually consistent with the real world. To fill the gap, we propose a highly credible and controllable generative pre-training technique (see Figure.~\ref{fig: control-generations}), in which a self-supervised adversarial loss and a controllable language modeling loss are optimized during the pre-training phase. In detail, a self-supervised adversarial loss allows the model to learn to distinguish whether a text is the original one or generated by ERNIE 3.0. Besides accelerating the convergence of the model, this loss enables ERNIE 3.0 Titan to re-ranking the credibility of the generated results. Meanwhile, a controllable language modeling loss is applied to enable the model to generate texts with specific attributes. We prompt the original text with a diverse attribute set, including the genre, topic, keywords, sentiment, and length, which can be easily expanded for more user-defined attributes. The users can freely combine these attributes for controllable generations of different types of downstream application scenarios. We conduct several experiments on 68 datasets. The results show that ERNIE 3.0 Titan significantly outperforms previous models on various tasks by a large margin and achieves new state-of-the-art results.
%In addition, attributes such as genre also improve the performance of ERNIE 3.0 Titan by better aligning the model within domain-specific tasks.

Furthermore, we propose a distillation framework to distill the ERNIE 3.0 Titan for easy deployment. Intuitively, we can apply current knowledge distillation methods~\cite{tinybert,minilm,minilmv2,ernie-tiny} to ERNIE 3.0 Titan. However, current distillation methods require an additional inference stage on a fully trained teacher to transfer knowledge to the student models, which is not environment-friendly concerning carbon emissions~\cite{carbon}. Another problem of current methods is that only one student model can be produced after the distillation phase is completed, requiring the teacher to infer multiple times to distill multiple students. In addition to the computation resource problems, previous studies ~\cite{teacher-assistant, KDefficacy, rco, follow-your-path} reveal that distillation from oversized teachers can lead to unexpected performance degradation problems.
~\cite{teacher-assistant, KDefficacy, skd} indicates that the difficulty comes from the large gap between the teacher's and student's parameter numbers, causing significant differences between their representation spaces. To this end, we propose an online distillation framework to efficiently distill the ERNIE 3.0 Titan into multiple small models during the pre-training stage, which results in little additional computation cost as compared to current distillation methods. Our distillation framework contains four key features: i)  teaching multiple students at the same time, ii) proposing On-the-Fly Distillation (OFD), where the teacher instructs the students during the training stage for a more environmentally friendly distillation, iii) introducing teacher assistants~\cite{teacher-assistant} for better distilling large scale models, iv) introducing Auxiliary Layer Distillation (ALD), a technique to improve distillation performance by stacking an additional student layer in distillation stage and discarding it at the fine-tuning stage. We compare our distilled ERNIE 3.0 Titan with previous compact models on five typical types of downstream tasks. The results demonstrate the effectiveness of our proposed distillation framework, and show that the distilled ERNIE 3.0 Titan achieves SOTA results on all tasks.

\section{Related Work} 
\subsection{Large-scale Pre-training}

Due to the rapid development of deep learning algorithms and the iterations of high-performance chips, pre-trained language models such as BERT~\cite{devlin2018bert}, GPT~\cite{radford2018improving}, and ERNIE~\cite{sun2019ernie} have made significant breakthroughs in many fields of natural language processing, such as natural language understanding, language generation, machine translation, human-machine dialogue, and question answering. These methods use unified Transformer models for learning universal representations on large unsupervised corpora. This technique taps into the advantages of scale of unsupervised data brings to natural language processing and significantly breaks the high reliance on costly annotated data. 

Some recent works~\cite{Megatron-LM,T5,gpt-3,switch-transformer} had demonstrated that increasing the size of pre-trained models can further exploit the potential value of unsupervised data. For example, the T5 model~\cite{T5} was proposed to push the performance for both natural language understanding and natural language generation tasks with 11 billion parameters. The T5 model converts all text-based language tasks into a text-to-text format by a unified framework and fully explores the effectiveness of pre-training objectives, architectures, unlabeled datasets, transfer approaches, and other factors. GPT-3~\cite{gpt-3}, with 175 billion parameters, achieved an amazing performance on a wide range of tasks under the few-shot and zero-shot settings with prompts. Several work have investigated the effects of larger pre-trained models such as Jurassic-1~\cite{jurassic-1},  Gopher~\cite{gopher}, Megatron-Turing NLG~\cite{megatron-nlg}, PanGu-$\alpha$~\cite{zeng2021pangu}, Yuan 1.0~\cite{wu2021yuan}, etc. An alternative route to increasing the size of the model parameters is to sparse the model parameters. These models~\cite{switch-transformer,lewis2021base,roller2021hash,zhang2021cpm,lin2021m6} use mixture-of-experts to select different parameters for each incoming example. As far as we know, sparse models have not achieved better performance than dense models up to now, although they can scale more efficiently with trillion parameters or even more. Therefore, this paper mainly focuses on large-scale distributed training and inference techniques for dense models. Most of the previous models learned only on plain texts confronted the problem of lacking common sense~\cite{rise_risk_gpt3}. In addition, most large-scale models are trained in an auto-regressive way, but~\cite{devlin2018bert} shows that such models have poorer performance with traditional fine-tuning when adapting to downstream language understanding tasks. In order to solve these problems, a unified framework called ERNIE 3.0~\cite{sun2021ernie} was proposed to train large-scale knowledge-enhanced models on large-scale plain texts and a large-scale knowledge graph by fusing the auto-regressive network and the auto-encoding network.

The exponential increment of the pre-trained language model's size has posed a great challenge for efficient training due to memory constraints and unaffordable training time. The size of pre-trained language models exceeds the memory limit of a single modern processor. In addition,  it is inevitable to store momentum and other optimizer states in widely used optimization algorithms such as Adam~\cite{kingma2014adam}. Therefore there are lots of work focusing on achieving efficient training of large-scale models. The first category is the Pipeline Model Parallelism, splitting different Transformer layers into separate cards. GPipe~\cite{huangGPipeEfficientTraining2019} utilizes a novel batch-splitting pipelining algorithm, resulting in almost linear speedup when partition model across multiple accelerators. TeraPipe~\cite{liTeraPipeTokenLevelPipeline2021} proposed a high-performance token-level pipeline parallel algorithm for synchronous model-parallel training of uni-directional language models. PTD-P~\cite{narayanan2021efficient} scheduled with interleaved stages in a fine-grained way to reduce the size of the pipeline bubble further.  Another category is the Tensor Model Parallelism~\cite{Megatron-LM}, in which individual layers of the model are partitioned over multiple devices. They took advantage of the structure of transformer networks to create a simple model parallel implementation by adding a few synchronization primitives. PTD-P~\cite{narayanan2021efficient} also combine pipeline, tensor, and data parallelism across multi-GPU servers to combine their advantages.

\subsection{Credible Text Generation}
The credibility of a text contains multiple aspects such as coherency, clarity, veracity, etc. For non-pretraining methods, previous works explored many approaches to promote several aspects of credible generation: Plan-and-write~\cite{yao2019plan} uses a hierarchical method that first plans a storyline and then generates a story based on it to improve the text coherency; CGRG~\cite{wu2020controllable} uses grounding knowledge to generate veritable answers. With the development of pre-training technology, large-scale language models provide a simple yet powerful solution for credible text generation. ~\cite{kreps2020all} shows that GPT-2~\cite{radford2019language} synthetic text samples can achieve a promising convincing score compared with authentic articles from the New York Times  (72\% vs. 83\% in one cohort judged the articles to be credible). Furthermore, GROVER~\cite{zellers2019defending} utilizes an adversarial framework to train a classifier with the generated text for incredible news detection. Inspired by this, ERNIE 3.0 Titan introduces an auxiliary adversarial loss to train a credibility ranker for self-ranking and select the most credible text for output.

\subsection{Controllable Text Generation}
As large-scale pre-training models have shown growing effectiveness in generating high-quality sentences, controllable text generation is getting more attention~\cite{NitishShirishKeskar2019CTRLAC}. GPT-3~\cite{gpt-3} used in-context few-shot learning to prompt text generation for various tasks, while it is still challenging for users in controlling the generation results. CTRL~\cite{NitishShirishKeskar2019CTRLAC} provided an effective way for controllable text generation. It trained a language model conditioned on control codes that govern style, content, and task-specific behavior. However, these control codes, derived from the structure that naturally co-occurs with raw texts, cover constrained controllable attributes. Following works like Arg-CTRL~\cite{BenjaminSchiller2020AspectControlledNA} and Tailor~\cite{AlexisRoss2021TailorGA} extended the control codes of CTRL either by crowdsourcing aspect annotations or deriving from the PropBank formalism to control the semantic-specific generation. Another line of work~\cite{SumanthDathathri2019PlugAP, AlvinChan2020CoConAS} aims to generate texts of desired attributes through relatively small ‘pluggable’ attribute models focusing on light-weight controllable fine-tuning on pre-trained models. In this paper, we focus on directly providing a highly controllable model pre-trained on ERNIE 3.0 controllable dataset powered by conditional LM loss.

\subsection{Model Distillation of Language Models}
Although large-scale language models show their outstanding ability on various tasks, their enormous parameters require a significant amount of computational resources and make them difficult to deploy on real-life applications. Language model distillation~\cite{tinybert,ernie-tiny,minilm,minilmv2,pkd,mobilebert,well-read,distill-bert,sixteen} has recently drawn significant attention as one prevailing method to solve the foregoing problem.
For example, \cite{distill-bert} proposes DistilBERT to successfully halve the depth of the BERT model by matching the output between teacher and student model on pre-training and fine-tuning stages. TinyBERT \cite{tinybert} adds two additional points of distillation, the attention distribution and hidden representations, to improve distillation quality. To further boost the distillation, in addition to the pre-training and fine-tuning stages, ERNIE-Tiny \cite{ernie-tiny} introduces two additional stages to ensure that the student captures the domain knowledge from the fine-tuned teacher. Unlike previous works that require a fine-tuned teacher for distillation, \cite{minilm} proposes task-agnostic distillation that only performs distillation on the pre-trained teacher by mimicking self-attention and value relation. 

Recent works~\cite{teacher-assistant, KDefficacy, rco, skd, akd} have observed that distillation from an oversized teacher to a significantly smaller student can cause an unexpected performance drop. \cite{teacher-assistant} suggests that the performance drop is due to the capacity gap between the teacher and the student, and introduces additional student models named teacher assistants with size between teacher and student to alleviate this gap. \cite{KDefficacy} proposes to early stop the teacher to reduce the gap. \cite{rco} proposes to utilize unconverged teacher checkpoints to guide student's learning process in a curriculum learning manner. \cite{follow-your-path} proposes joint training of teacher and student, allowing the teacher to be aware of the existence of the student and reducing the gap, although it deteriorates the performance of the teacher model. \cite{skd} suggests that the unexpected distillation performance drop is caused by the fact that the oversized model tends to be overconfident, and forcing the student to learn such overconfident output harms the distillation performance. It proposes to normalize the point of distillation to alleviate the overconfidence problem.  

\section{ERNIE 3.0 Titan Model} 
 
 \begin{figure*}[ht]
	\centering
	\includegraphics[width=1.0\textwidth]{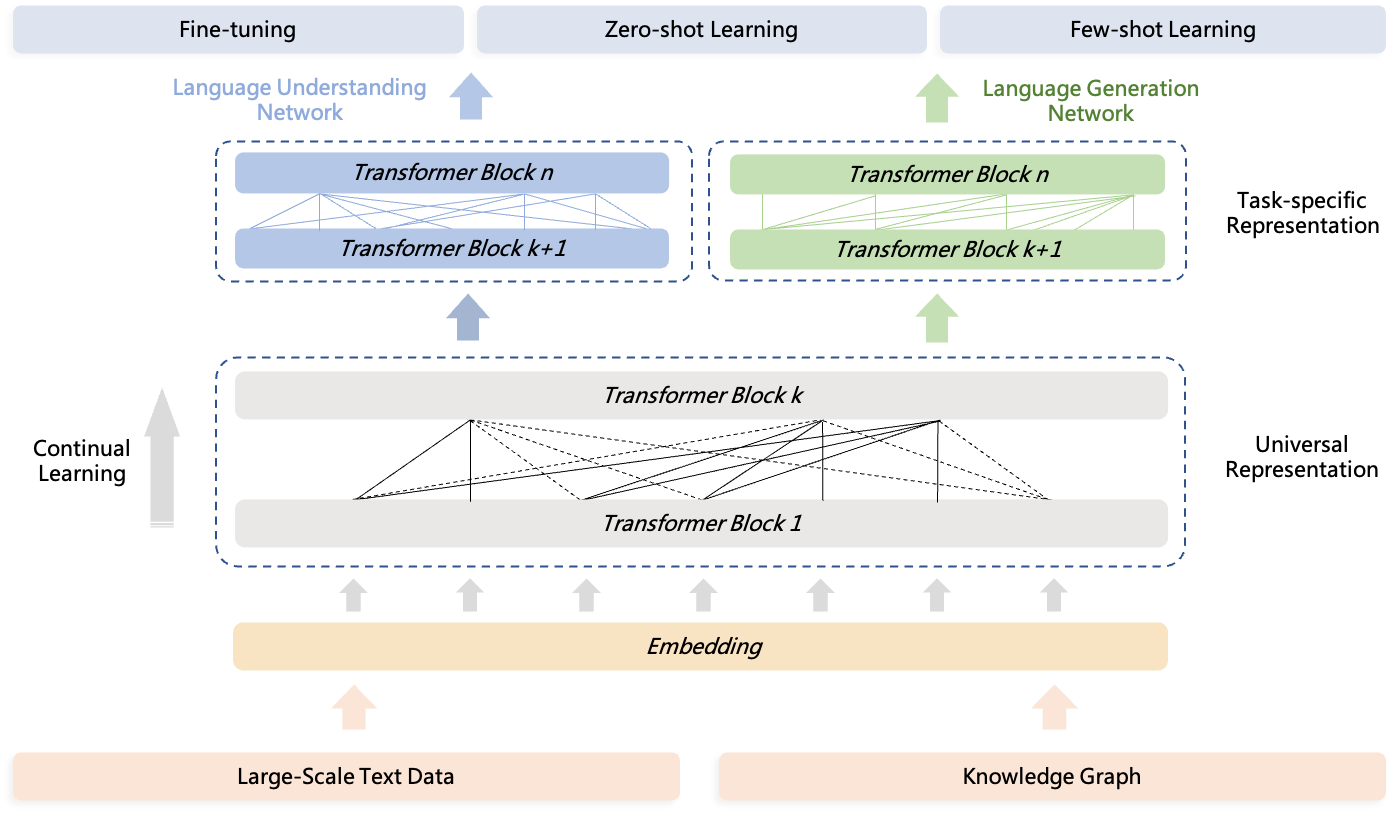}
	\caption{The framework of ERNIE 3.0.} \label{label:ernie 3.0 framework}
\end{figure*}

A significant improvement has been achieved on various natural language processing tasks for knowledge-enhanced pre-trained models with the base or large model size, such as ERNIE, ERNIE 2.0, and SpanBERT~\cite{joshi2020spanbert}, in which the base/large model size represent 12/24 layers Transformer respectively. In order to explore the effectiveness of knowledge enhanced large-scale pre-trained model, a \textbf{\textit{Continual Multi-Paradigms Unified Pre-training Framework}} named ERNIE 3.0 Framework is proposed in \cite{sun2021ernie} to pre-train model on massive unsupervised corpus including plain texts and knowledge graphs. Specifically, ERNIE 3.0 Framework allows collaborative pre-training among multi-task paradigms, in which various types of pre-training tasks are incrementally deployed in the corresponding task paradigm to enable the model to learn different levels of knowledge, i.e., valuable lexical, syntactic and semantic information, more effectively. Benefiting from the superiority of ERNIE 3.0 Framework, ERNIE 3.0 has made astonishing improvements on abundant downstream tasks across natural language understanding and natural language generation. As a matter of course, ERNIE 3.0 Titan is built on ERNIE 3.0 Framework in this paper. The detail of ERNIE 3.0 Framework will be explained in the following sections.

\subsection{Overview of ERNIE 3.0 Framework}
Until recently, the prevalent unified pre-training models trend to employ a shared Transformer network for different well-designed cloze tasks and utilize specific self-attention masks to control the context of the prediction conditions. Nevertheless, 
we believe that the different task paradigms of natural language processing consistently depend on identical underlying abstract features, such as lexical and syntactic information. However, the requirements of top-level concrete features are incompatible, in that the natural language understanding tasks have the disposition to learn the semantic coherence while natural language generation tasks expect further contextual information. Inspired by the classical model architecture of multi-task learning, in which the lower layers are shared across all tasks while the top layers are task-specific, \cite{sun2021ernie} construct the ERNIE 3.0 Framework shown in Figure \ref{label:ernie 3.0 framework}, which enable the different task paradigms to share the underlying abstract features learned in a shared network and utilizing the task-specific top-level concrete features learned in their task-specific network respectively. The backbone shared network and task-specific networks are referred to as the \textbf{\textit{Universal Representation Module}} and \textbf{\textit{Task-specific Representation Module}}s in ERNIE 3.0 Framework. Specifically, the universal representation network plays the role of a universal semantic features extractor (for example, a multi-layer Transformer). The parameters are shared across all kinds of task paradigms, including natural language understanding and natural language generation, and so on. And the task-specific representation networks undertake the function of extracting the task-specific semantic features, in which the parameters are learned by task-specific objectives. Furthermore, the continual multi-task learning framework introduced in ERNIE 2.0~\cite{sun2020ernie} is integrated into ERNIE 3.0 Framework to help the model efficiently learn the lexical, syntactic, and semantic representations.

Driven by the success of ERNIE 3.0 \cite{sun2021ernie}, ERNIE 3.0 Titan also employs the collaborative architecture of a Universal Representation Module and two Task-specific Representation Modules, namely natural language understanding (NLU) specific representation module and natural language generation (NLG) specific representation module. Details are as follows: 
\begin{itemize}
    \item \textbf{\textit{Universal Representation Module}}. In likewise, a multi-layer Transformer-XL \cite{dai2019transformer} is adopted as the backbone network like other pre-trained models such as ERNIE 3.0 \cite{sun2021ernie}, XLNet \cite{yang2019xlnet}, Segatron \cite{DBLP:journals/corr/abs-2004-14996} and ERNIE-Doc \cite{ding2020ernie}, in which Transformer-XL is similar to Transformer but introduces an auxiliary recurrence memory module to help modelling longer texts. Proverbially, the larger the scale of the Transformer model, the stronger its capacity to capture and store up various semantic information with different levels enabled by the self-attention mechanism. Therefore, ERNIE 3.0 Titan sets the universal representation module with a larger size (refer to the section \ref{pretraining_setting}) to enable the model to effectively capture universal lexical and syntactic information from training data by learning various pre-training tasks of different paradigms. And what needs special attention is that the memory module is only valid for natural language generation tasks while controlling the attention mask matrices. 
    \item \textbf{\textit{Task-specific Representation Modules}}. Instead of the multi-layer perceptron or shallow Transformer commonly used as task-specific representation networks in multi-task learning, ERNIE 3.0 Titan employs the Transformer-XL network with base model size as the task-specific representation modules to capture the top-level semantic representations for different task paradigms. Under this design, ERNIE 3.0 Titan achieves a triple-win scenario: the base Transformer has a stronger ability to capture semantic information than multi-layer perceptron and shallow Transformer while not significantly increasing the parameters of the large-scale model; last but not least, a new available route that enables the realizable practical applications for large scale pre-trained model can be explored --- only fine-tuning on the task-specific representation modules. ERNIE 3.0 Titan constructs two task-specific representation modules: the bi-directional modeling NLU-specific representation network and 
    the uni-directional modeling NLG-specific representation network.
\end{itemize}

\subsection{Pre-training Tasks}
\label{sec:pre-training tasks}

In order to make the capacity of understanding, generation and reasoning available to ERNIE 3.0 Titan, we construct several tasks for various task paradigms to capture different aspects of information in the training corpora, including word-aware pre-training tasks, structure-aware pre-training tasks and knowledge-aware pre-training task introuced in ERNIE 3.0 \cite{sun2021ernie}. Additionally, an innovative 
knowledge-aware pre-training task namely \textbf{\textit{Credible and Controllable Generations}} is built to control the generation result and obtain the result factually consistent with the real world.

\subsubsection{Word-aware Pre-training Tasks}\label{sec:Word-aware pretrain-task}

\textbf{Knowledge Masked Language Modeling} \,\, 
ERNIE 1.0~\cite{sun2019ernie} proposed an effective strategy to enhance representation through knowledge integration, namely Knowledge Integrated Masked Language Modeling task. It introduced phrase masking and named entity masking that predict the whole masked phrases and named entities to help the model learn the dependency information in both local contexts and global contexts. 

\textbf{Document Language Modeling} \,\, 
As introduced in \cite{sun2021ernie}, document language modeling task is a special version of traditional language modeling task, which trains models on long text instead of the prevailing shorter segments of manageable size (at most 512 tokens). Enhanced Recurrence Memory Mechanism proposed in ERNIE-Doc~\cite{ding2020ernie} is introduced into ERNIE 3.0 Titan to heighten the capability of modeling a larger effective context length than traditional recurrence Transformer. 

\subsubsection{Structure-aware Pre-training Tasks}\label{sec:Structure-aware pretrain-task}
\textbf{Sentence Reordering} \,\, 
Sentence reordering task, which is introduced in ERNIE 2.0~\cite{sun2020ernie}, aims to train the model to learn the relationship between sentences by reorganizing permuted segments. At length, a given paragraph is randomly split into 1 to m segments during pre-training, and all of the combinations are shuffled by a randomly permuted order. Then, the pre-trained model is asked to reorganize these permuted segments, modeled as a k-class classification problem where \begin{math} k=\sum_{n=1}^{m} n!\end{math}.

\textbf{Sentence Distance} \,\, 
Sentence distance task, an extension of the traditional next sentence prediction (NSP) task, is widely used in various pre-trained models to enhance their ability to learn the sentence-level information, which can be modeled as a 3-class classification problem. The three categories represent that the two sentences are adjacent, nonadjacent but in the same document and from two different documents, respectively.

\subsubsection{Knowledge-aware Pre-training Task}\label{sec: pretrain-task}
\textbf{Universal Knowledge-Text Prediction} \,\,
Universal knowledge-text prediction (UKTP) task, a particular masked language modeling that constructed on both unstructured texts and structured knowledge graphs, plays a pivotal role in incorporating world knowledge and commonsense knowledge into pre-trained model. Given a pair of a triple from the knowledge graph and the corresponding sentence from the encyclopedia, UKTP task randomly mask the relation in triple or the words in corresponding sentence. To predict the relation in the triple, the model needs to detect mentions of the head entity and the tail entity and determine their semantic relationship in the corresponding sentence. Another, to predict the words in the corresponding sentence, the model needs to consider the dependency information in the sentence and the logical relationship in the triple.

\textbf{Credible and Controllable Generations} \,\, 
 \begin{figure*}[ht]
	\centering
·   	\includegraphics[width=1.0\textwidth]{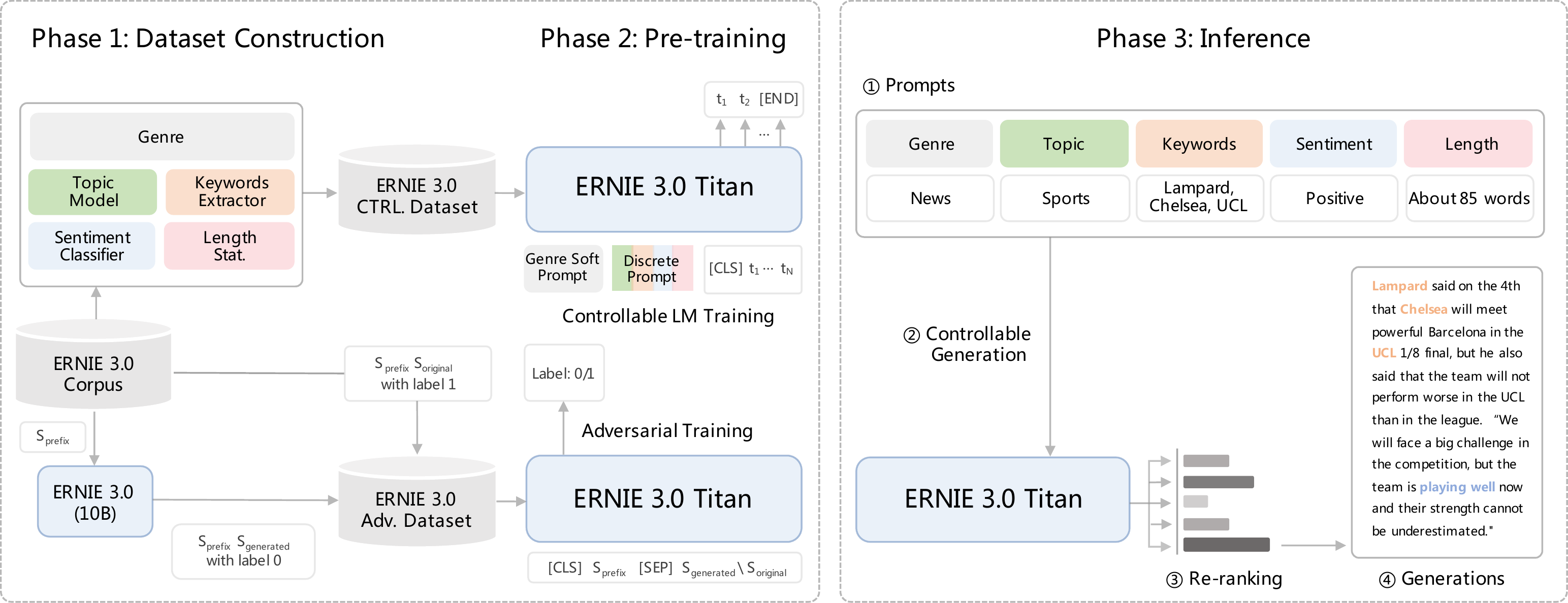}
	\caption{The framework of ERNIE 3.0 Titan on credible and controllable generations.} \label{fig: control-generations}
\end{figure*}
Controlling the generated texts based on desired attributes and improving their credibility is a key and practical feature we introduced in ERNIE 3.0 Titan. To achieve this, we design a self-supervised adversarial loss and a controllable language modeling loss for generating credible and controllable texts, respectively. 

The self-supervised adversarial loss allows the model to distinguish whether a text is generated or the original one. As a result, it is easy for ERNIE 3.0 Titan to discard the low credibility generated texts with repeating words, unfluent and conflicting sentences. In detail, we formalize this as a binary classification problem experimented on our ERNIE 3.0 adversarial dataset $D_{a} = \{D_{\text{original}}, D_{\text{generated}}\}$ which is a subset of original ERNIE 3.0 Corpus ${D_{\text{original}}}$ with its adversarial samples $D_{\text{generated}}$ generated by ERNIE 3.0.

\begin{equation}
    \mathcal{L}_a(D_{a}) = - \sum_{n=1}^{|D_{a}|} \log P_{\theta}(y^n=\mathbb{I}_{\mathbf{h}_{\textit{[CLS]}}^n \in D_{\text{original}}}| \mathbf{h}_{\textbf{[CLS]}}^n)
\end{equation}
where we trained ERNIE 3.0 Titan with parameters $\theta$ to minimize the cross-entropy loss, $\mathbb{I}_{\mathbf{h}_{\text{[CLS]}}^n \in D_{\text{original}}}$ indicates whether $n_{\text{th}}$ sample belongs to the original dataset $D_{\text{original}}$, and the output hidden state of the special token $\textit{[CLS]}$ is taken as input for binary classification.

The controllable language modeling loss is a modified language modeling loss by conditioning on extra prompts for controlling the generated texts as follows:

\begin{eqnarray} 
\mathcal{L}_{c}(D_{c}) = \begin{cases}
-\sum_{n=1}^{|D_{c}|} \log{P_\theta(x_t^n|x^n_{<t})} & \text{if prob.} p \le  0.5\\
-\sum_{n=1}^{|D_{c}|} \log{P_\theta(x_t^n|x^n_{<t}, \text{prompts}^{n}}) & \text{if prob.} p > 0.5\\
\end{cases}
\label{eq:controllable_loss}
\end{eqnarray}
where ERNIE 3.0 Titan is trained to minimize the negative log-likelihood loss on ERNIE 3.0 controllable dataset $D_{c}=\{x^1, x^2, \dots, x^{|D_{c}|}\}$, $t$ means the $t_{\text{th}}$ token of $x$. $x^n$ is associated with $\text{prompts}^{n}$ specifying the genre, topic, keywords, sentiment and length. The loss is switched to the normal language modeling loss with a pre-defined probability 0.5 to prevent the model from heavily depending on the prompts. Different from CTRL which convers a constrainted controllable attributes from the semi-structural raw texts, we enrich the controllable attributes set using task-specific supervised models on ERNIE 3.0 Corpus. As the ERNIE 3.0 Corpus in nature constructed from various sources including Web, QA, Novel, Knowledge graph and etc. (see ~\ref{sec:pre-training data}), we assign soft prompts (learnable prompt embedding) for different datasets to better align the model within the genre of the target dataset.

\subsection{Pre-training Data}\label{sec:pre-training data} 
To ensure the success of the pre-training of ERNIE 3.0 Titan, we utilize the ERNIE 3.0 Corpus~\cite{sun2021ernie}, a large-scale, wide-variety, and high-quality Chinese text corpora amounting to 4TB storage size in 11 different categories. Two additional datasets, namely the ERNIE 3.0 adversarial dataset and ERNIE 3.0 controllable dataset, are constructed.

\noindent{\textbf{ERNIE 3.0 adversarial dataset}}: The adversarial dataset is constructed based on ERNIE 3.0 Corpus. The positive examples consist of 2M natural paragraphs sampled from ERNIE 3.0 Corpus, while for negative example generation, we randomly take the first 1\textasciitilde 3 sentences from the original positive paragraph as the prefix input, and utilize ERNIE 3.0 to generate the rest part of the paragraph. The  max length of generated paragraph is set to 512, and we discard the last incomplete sentence if the generation process is terminated by max-length excess.

\noindent{\textbf{ERNIE 3.0 controllable dataset}}: The controllable dataset is highly scalable to include more diverse user-defined attributes. Here, we introduce 5 different controllable attributes including genre, topic, keywords, sentiment and length as follows:
\begin{itemize}
    \item \textbf{Genre} is assigned to samples based on the source the data collected from, including general (ERNIE 2.0 Corpus), Web, QA, Knowledge, Finance, Law, Medical, Novel, Couplet, Poet, Machine Translation. Each genre type is associated with pre-defined maximum soft prompt embeddings (64 in our experiment). In the pre-training phase, the number of soft prompt embeddings are sampled randomly between 0 and the maximum number.
    \item \textbf{Topic} is labeled using a topic model~\footnote{\url{https://ai.baidu.com/tech/nlp_apply/topictagger}} which can classify a document into 26 different topics such as international, sports, society, news, technology, digital, emotion, cars, education, fashion, games, travel, food, culture, healthy life, child and music.
    \item \textbf{Keywords} are extracted using a keyword extraction model~\footnote{\url{https://ai.baidu.com/tech/nlp_apply/doctagger}} which performs in-depth analysis of article titles and content, and outputs multi-dimensional tags that reflect the key information of the article, such as subject, entity, etc.
    \item \textbf{Sentiment} is derived using a sentiment classification model~\footnote{\url{https://ai.baidu.com/tech/nlp_apply/sentiment_classify}}. A positive, negative, or neutral label is assigned to each sample.
    \item \textbf{Length} is counted on the tokenized text. The length attribute can prompt the model to generate texts with the desired length to avoid harshly truncating.
\end{itemize}

In pre-training phase, we use the following input format for each sample:  ``\texttt{[Genre-0], [Genre-1], $\cdots$ [Genere-N] [t] Topic texts [/t] [k] Keyword text0, Keyword text1, $\cdots$ [/k] [senti] Sentiment label text [/senti] [w] About $L$ words [/w] Original text} '' where $\texttt{[Genere-$n$]}$ is $n_{\text{th}}$ soft prompt embedding for one of the genre type mentioned above, $N \in [0, 64)$, $L$ is the token number of the original text and \texttt{[t], [/t], [k], [/k] ,[senti], [/senti], [w], [/w]} are special tokens to seperate each attribute. For example, the input for the case in Figure.~\ref{fig: control-generations} can be ``\texttt{[News-0], [News-1], $\cdots$ [News-N] [t] Sports [/t] [k] Lampard, Chelsea, UCL [/k] [senti] Positive [/senti] [w] About 85 words [/w] Lampard said on the 4th that Chelsea...} ''. Note that for each attribute, we randomly discard it with a pre-defined probability (0.5 in our experiment) to prevent the model from heavily depending on it.
 
\subsection{Pre-training Settings} \label{pretraining_setting}
Following the pre-training setting of ERNIE 3.0, ERNIE 3.0 Titan includes the universal representation module and the task-specific representation modules, which both use the Transformer-XL structure. 
We adopt a structure with 12 layers, 768 hidden units, and 12 heads for the task-specific representation modules. 
We adopt a structure with 48 layers, 12288 hidden units, and 192 heads for the universal representation modules.
We found that continually increasing the hidden size would make it difficult to load the parameter of output embedding in a single machine with eight cards with 32GB memory. In order to further increase the model capacity, we choose to scale the parameters of the point-wise feed-forward network alternatively. The inner layer of the universal representation modules has a dimensional of 196608, which is 16 times the hidden size of the model.
The total number of parameters of ERNIE 3.0 Titan is over 260 billion.

We use Adam~\cite{kingma2014adam} with learning rate of 1e-4, $\beta_1=0.9$, $\beta_2=0.95$, L2 weight decay of 0.1, We also clip the global norm of the gradient at 1.0 to improve the stability of pre-training. The maximum sequence length of context and the memory length of language generation is 512 and 128, respectively. We use progressive learning to speed up convergence in the first 4000 steps and linear decay of the learning rate. ERNIE 3.0 Titan is implemented on the PaddlePaddle framework and uses parallelism techniques that facilitate the efficient training of large models that do not fit in the memory of a single GPU. We will give the detail of these in the next section.

\setcounter{secnumdepth}{4}
\section{Efficient Training and Inference of ERNIE 3.0 Titan} 

\subsection{Distributed Training}
To train a huge model like GPT-3 faces severe challenges in memory consumption and computational efficiency. This model requires 2.1TB for parameter and optimizer states storage and $3.14E^{11}$ TeraFLOPS for training 300 billion tokens. Given that a modern AI processor like Nvidia V100 GPU can only provide 32GB of memory and 125 TeraFLOPS~\footnote{\url{https://images.nvidia.com/content/volta-architecture/pdf/volta-architecture-whitepaper.pdf}}, it will take 28 days to train with 2048 GPU V100 cards even with a 50\% percentage of theoretical peak FLOPS. There have been many related works to solve these problems, for example, Megatron-LM~\cite{megatron}, Gpipe~\cite{huangGPipeEfficientTraining2019}, Zero~\cite{rajbhandariZeROMemoryOptimizations2020} , and so on. PaddlePaddle~\cite{ma2019paddlepaddle} has also proposed 4D hybrid parallelism with more sophisticated combination techniques~\footnote{\url{https://ai.baidu.com/forum/topic/show/987996}}. 

To train the ERNIE 3.0 Titan model on heterogeneous clusters is faced with even more challenges than GPT-3.
On the one hand, the ERNIE 3.0 model has a more sophisticated architecture than GPT-3, such as task-specific representation layers and memory mechanism~\cite{DBLP:journals/corr/abs-2012-15688}. Such structures are prone to unbalance and inefficient pipeline training on top of intensive memory consumption. On the other hand, different clusters are coupled with distinct software stacks, including operator implementations, communication libraries, schedulers, etc. For example, NPU performs worse than GPU on the tensor computation of dynamic shapes and shows strength on FP16 calculation, posing new challenges like customized 
layers optimization, unbalanced pipeline, and training stability.
 
To achieve efficient and stable training with convergence guarantee, PaddlePaddle developed an end-to-end adaptive distributed training technology, including fine-grained parallelism, heterogeneous hardware-aware training, and fault tolerance mechanism to train the 260B model on both Nvidia V100 GPU and Ascend 910 NPU clusters.

\subsubsection{Fine-grained Hybrid Parallelism}
The 4D hybrid parallelism includes data parallelism (DP)~\cite{horovod, liPyTorchDistributedExperiences2020a}, intra-layer tensor model parallelism (MP) \cite{tensorflowmesh, megatron}, inter-layer pipeline model parallelism (PP) \cite{huangGPipeEfficientTraining2019, pipedream, pipedream2bw, narayananMemoryEfficientPipelineParallelDNN2021, liTeraPipeTokenLevelPipeline2021}, and the improved sharded data parallelism (Sharded) based on ZeRO \cite{rajbhandariZeROMemoryOptimizations2020}. DP is deployed to partition and distribute the data across devices and synchronize all their gradient. PP is utilized to split model layers into multiple stages distributed across devices, where the throughput is directly related to load balancing and bubble fraction. MP is used to slice parameters and activation as distributed across devices. Sharded commits to reducing the redundancy of optimizer states. Our improved version of Sharded, namely Group Sharded, is designed to decouple data-parallel and Sharded flexibly.

Besides, we implemented distributed operators for large Embedding, FC, and  FC\_Softmax layers, supporting fine-grained partition. These operators can utilize intra-node communication to further reduce memory occupation and redundant calculations. This strategy increases the overall throughput significantly on the ERNIE 3.0 Titan model's unique architecture.

\begin{figure*}[!htb]
  \centering
  \includegraphics[width=0.90\textwidth]{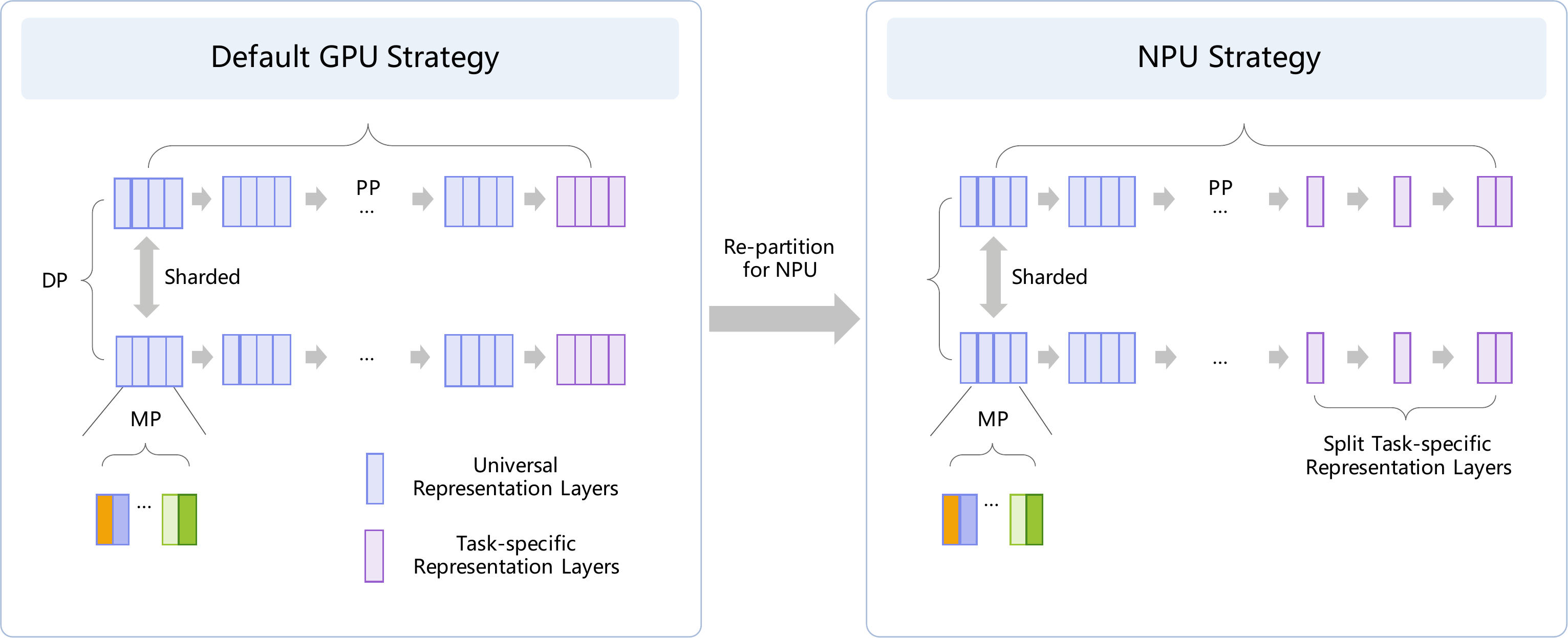}
  \caption{Resource-aware-partition.} 
  \label{fig:resource-aware}
\end{figure*}

\subsubsection{Heterogeneous hardware-aware training}
We train ERNIE 3.0 Titan model on both NPU clusters at PengCheng Lab and GPU clusters in Baidu. Specifically, training large-scale models on NPU relies heavily on carefully treating the following problems: 1) interface design between deep learning framework with fast-evolving NPU software stack; 2) software-hardware collaborative NPU performance optimization 3) adjustment of load balance concerning NPU cluster throughput. 

To take the best advantage of PaddlePaddle's imperative programming and the 4D parallelism technology, we use Ascend Computing Language (ACL)~\footnote{\url{https://support.huaweicloud.com/asdevg-python-cann/atlaspython_01_0008.html}} in our implementation instead of using Ascend Graph(GE)~\footnote{\url{https://support.huawei.com/enterprise/en/doc/EDOC1100164817/753b4f6/introduction}}. We found that NPU has strong performance using pure FP16 while showing weakness in dealing with smaller shapes and dynamic shapes commonly used in NLP models.

\begin{figure*}[!htb]
  \centering
  \includegraphics[width=0.80\textwidth]{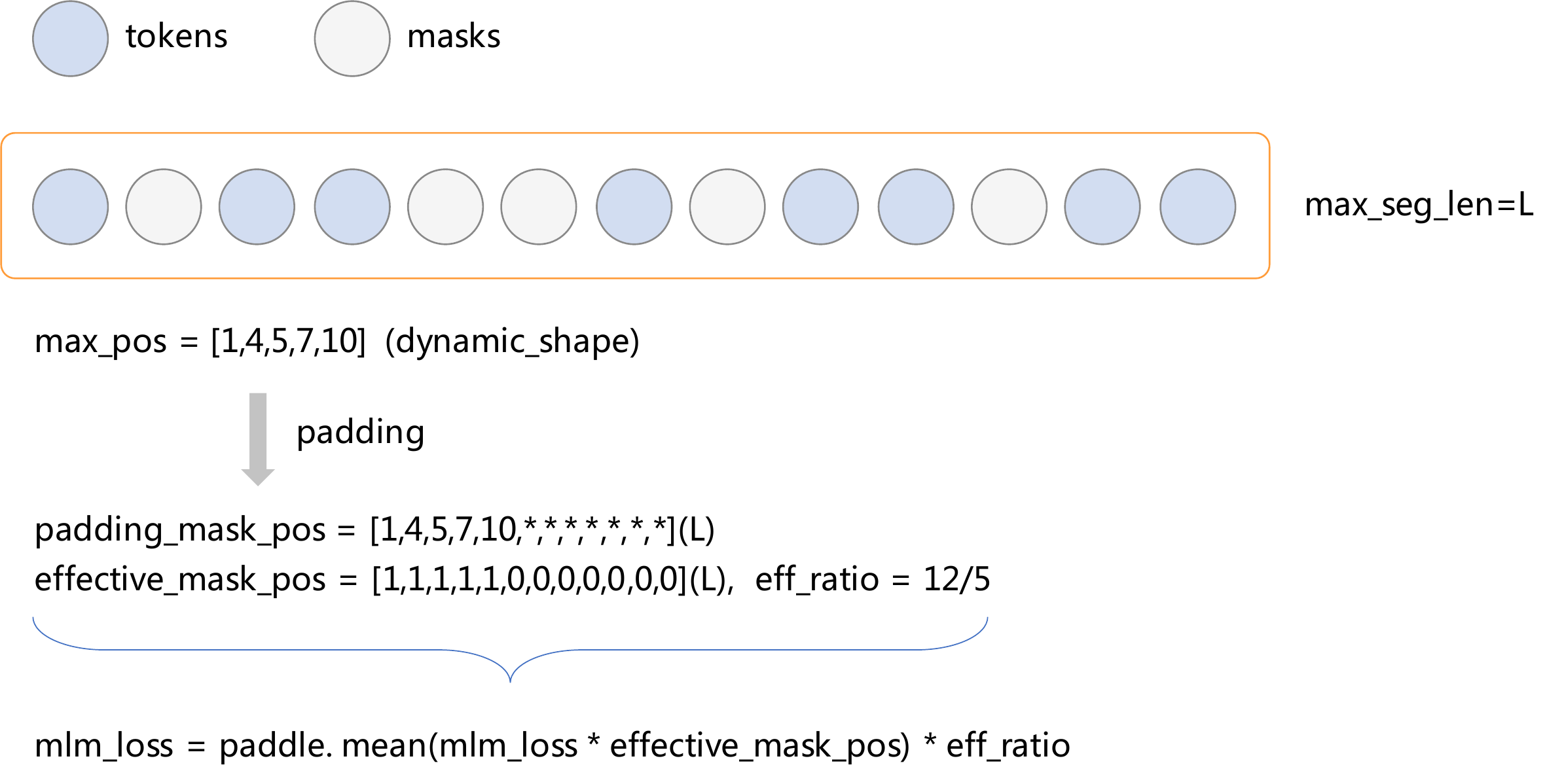}
  \caption{Convert dynamic shape to static shape.} 
  \label{fig:static-shape}
\end{figure*}

In the PaddlePaddle framework, the parallel strategy can be adjusted in a resource-ware manner to fully exploit the computing power of the hardware. Figure~\ref{fig:resource-aware} shows that the Task-specific Representation layers of ERNIE 3.0 require more load balance adjustments on NPUs than that on GPUs because NPUs consume more time on small layers' kernel launches. As shown in Table~\ref{tab:npu260B}, the ERNIE 3.0 Titan model reaches 91.7\% weak scalability with thousands of NPU cards and increases the throughput up to 2.1 times while the number of cards only increased by 22\%. Furthermore,  Figure~\ref{fig:static-shape} shows how we convert dynamic shape to static shape to utilize the ACL performance fully.

\subsubsection{Fault tolerance}
During our experiments, we also encountered hardware problems, for example, a sudden drop of communication speed because of PCIe errors, GPU card errors such as Double Bit ECC Error (DBE)~\footnote{\url{https://docs.nvidia.com/deploy/dynamic-page-retirement/index.html}}. PaddlePaddle developed fault-tolerant features that automatically replace faulty machines to reduce hardware waste and time consumption to resolve these problems. PaddlePaddle allows users to customize their need to store and restore states from the checkpoints.

\begin{table}[htb]
  \centering
  \caption{Comparison between the default configuration and the resource-aware configuration \\
        when training ERNIE 3.0 Titan on Ascend NPUs.}
  \label{tab:npu260B}
  \resizebox{0.6\textwidth}{!}{
  \begin{tabular}{c c c c c}
    \toprule  
    \makecell{\textbf{Configuration}} & 
    \makecell{\textbf{NPUs}}  & 
    \makecell{\textbf{DP}} &
    \makecell{\textbf{Global} \\ \textbf{batch size}} &
    \makecell{\textbf{Speedup}} \\
    \midrule
    \multirow{2}[1]{*}{\makecell{Default}} 
    & 392    & 1     & 512  & - \\
    & 1568    & 4     & 2048   & - \\
    \hline
    \multirow{2}[2]{*}{\makecell{Resource-aware-partition}} 
    & 480    & 1     & 512   & 2.19 \\
    & 1920    & 4     & 2048   & 2.17 \\
    \hline
    \bottomrule
  \end{tabular}}
\end{table} 

These problems encountered in practice prompt our work on end-to-end adaptive training. For details please refer to ~\cite{ao2021endtoend}.

\subsection{Distributed Inference}
It becomes infeasible to perform inference using ERNIE 3.0 Titan model on common GPU devices such as Nvidia A100-SXM4 (40GB). Therefore, the model has to be split into multiple subgraphs deployed on multiple GPU cards. We implemented tensor model parallelism and pipeline model parallelism in Paddle Serving, PaddlePaddle's online serving framework. In addition, we adopted methods such as unified memory optimization, op fusion, model IO optimization, and quantization-aware acceleration on a single card to improve the inference speed. 

\begin{figure*}[!htb]
	\centering
	\includegraphics[width=1.0\textwidth]{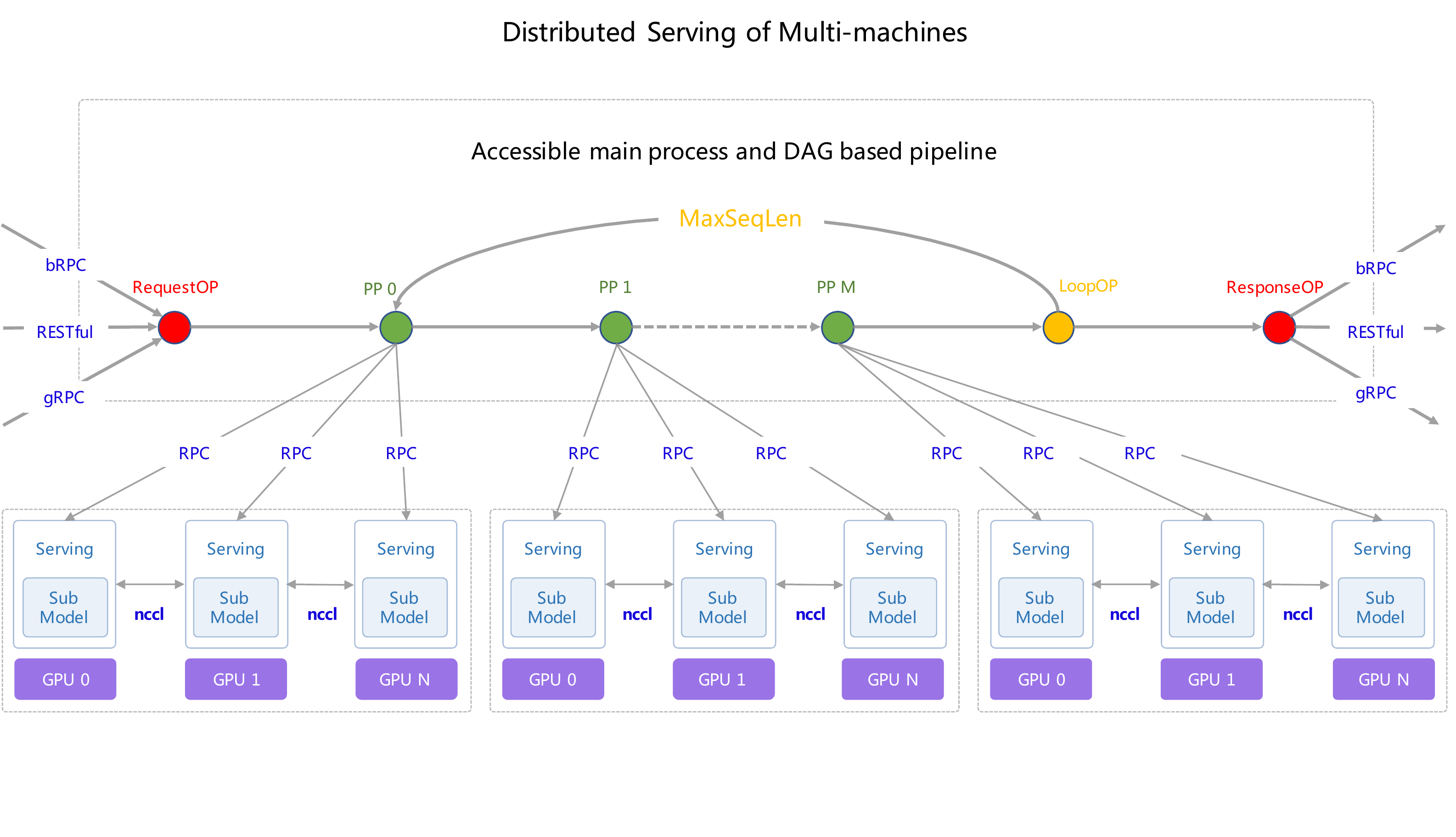}
	\caption{Distributed Inference framework of Paddle Serving.} 
\end{figure*}

\section{Online Distillation Framework for ERNIE 3.0 Titan}\label{sec:distill} 
We devise an online distillation framework concerning the computation overhead and carbon emissions during the training stage, where the teacher model will teach students and train itself simultaneously to utilize the computational resource more effectively. To describe the framework, we firstly introduce our main procedure and then discuss the two key techniques, \textbf{O}n the \textbf{F}ly \textbf{D}istillation (\textbf{OFD}) and \textbf{A}uxiliary \textbf{L}ayer \textbf{D}istillation (\textbf{ALD}), in detail.

\subsection{Main Framework of Distillation}
\begin{figure}[ht]
	\centering
	\includegraphics[width=0.8\textwidth]{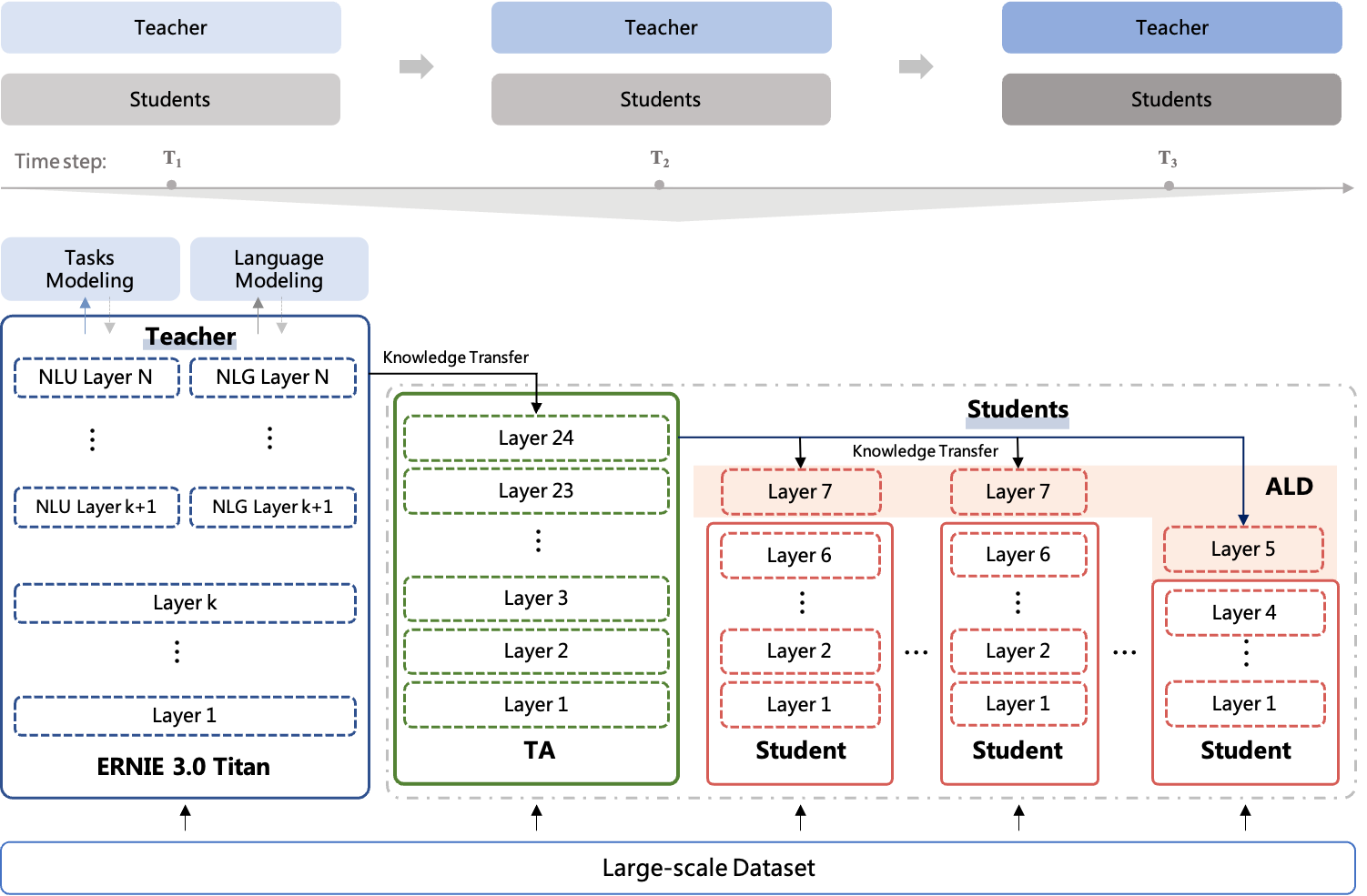}
	\caption{Online Distillation Framework for ERNIE 3.0 Titan.} 
	\label{compression}
\end{figure}

Unlike existing distillation methods, our proposed framework trains multiple students rather than one at once. Figure~\ref{compression} shows the general structure of our proposed distillation framework. The blue rectangle on the left represents the teacher model, ERNIE 3.0 Titan, the green rectangle represents a 24-layer model, which we call teacher assistant (TA). Besides the TA model, we also introduce multiple smaller student models (shown in red) into this framework. Considering the large gap between the teacher model and those smaller students, we do not directly transfer the knowledge from teacher to students. Instead, we use the teacher assistant \cite{teacher-assistant} as a distillation bridge to better transfer knowledge. As \cite{minilm,minilmv2} have shown that the attention probability of the teacher model's last few layers is crucial to task-agnostic knowledge distillation, we follow this paradigm in our framework. Through this framework, three students models can be trained simultaneously, saving the complex process to train them one by one in traditional distillation methods. However, there are still two issues with the current procedure. The first one is that this distillation process does not utilize the training time computational resource effectively and requires additional forward propagation from the teacher for distillation, causing additional computation overhead. The second problem is that matching attention module~\cite{minilm,minilmv2} in Transformer between teacher and students will leave the weights of feed forward network in the students' last layer untrained, as one Transformer layer is composed of an attention module and a feed forward network module~\cite{vaswani2017attention}. To solve those two problems, we introduce two techniques called OFD and ALD which will be discussed in the following sections.

\subsection{OFD: On the Fly Distillation} \label{sec:OFS}
To better utilize the computational resource and for a more environmentally friendly distillation, we propose our online learning method: OFD. The learning process is that every time the teacher updates one step, the students update one step toward the teacher. Students' learning targets (i.e. the teacher) change over time during the training process. From this "moving target" perspective, this framework is similar to ~\cite{rco} where different teacher checkpoints during pre-training are selected as distillation targets. However, the teacher in this framework changes smoothly, whereas that in ~\cite{rco} changes discretely. 
% TBD 
OFD allows teacher training and distillation perform simultaneously. 
The benefit is that we can better utilize the forward propagation from the teacher during the teacher's pre-training for distillation, unlike the existing knowledge distillation methods~\cite{tinybert,distill-bert} which requires additional forward propagation from the teacher to extract the knowledge for distillation. Note that the OFD will not influence the teacher's training as the gradients from distillation loss will not flow from the TA or students back to the teacher.  

\subsection{ALD: Auxilliary Layer Distillation} 
As the knowledge being transferred during distillation is the attention probability distribution, the other module in a Transformer block, the feed forward network, will not be trained during distillation. To show this problem more clearly, we will briefly describe the structure of the Transformer. 

In the Transformer architecture~\cite{vaswani2017attention}, each Transformer layer consists of two sub-modules, namely the multi-head self-attention ($\mathrm{MHA}$) and position-wise feed-forward network ($\mathrm{FFN}$). Transformer encodes contextual information for input tokens. The input embeddings $\{\mathbf{x}\}^{s}_{i=1}$ for sample $x$ are packed together into
$\mathbf{H}_{0}=\left[\mathbf{x}_{1}, \cdots, \mathbf{x}_{s}\right ] $ , where $s$ denotes the input sequence length. Then stacked Transformer blocks iteratively compute  the encoding vectors as $\mathbf{H}_{l}=\mathrm { Transformer }_{l}\left(\mathbf{H}_{l-1}\right), l \in[1, L]$, and the Transformer is computed as:

\begin{align}
\mathbf{A}_{l, a}&=\operatorname{MHA}_{l, a}(
\mathbf{H}_{l-1} \mathbf{W}_{l, a}^{Q}, \mathbf{H}_{l-1} \mathbf{W}_{l, a}^{K}), \label{mha-att}\\
\mathbf{H}_{l-1}' & = \text {LN}(\mathbf{H}_{l-1}+ (\mathop{\|}_{a=1}^h \mathbf{A}_{l, a}  (\mathbf{H}_{l-1} \mathbf{W}_{l, a}^{V})) \mathbf{W}_{l}^{O}) , \label{mha-res}\\
\mathbf{H}_{l}\  & = \text {LN}\left(\mathbf{H}_{l-1}'+\mathrm{FFN}\left(\mathbf{H}_{l-1}'\right)\right),
\label{mha-ffn}
\end{align}

where the previous layer’s output $\mathbf{H}_{l-1} \in \mathbb{R}^{ s \times d}$ is linearly projected to a triple of queries, keys and values using parameter matrices $\mathbf{W}^{Q}_{l,a}, \mathbf{W}^K_{l,a}, \mathbf{W}^{V}_{l,a} \in \mathbb{R}^{d \times d'}$, where $d$ denotes the hidden size of $\mathbf{H}_l$ and $d'$ denotes the hidden size of each head's dimension. $\mathbf{A}_{l,a} \in \mathbb{R}^{s \times s}$  indicates the attention distributions for the $a$-th head in layer $l$, which is computed by the scaled dot-product of queries and keys respectively. $h$ represents the number of self-attention heads. $ \| $ denotes concatenate operator along the head dimension. $\mathbf{W}_l^{O} \in \mathbb{R}^{d\times d}$ denotes the linear transformer for the output of the attention module. $\mathrm{FFN}$ is composed of two linear transformation function including mapping the hidden size of $\mathbf{H}_{l-1}'$ to $d_{ff}$ and then mapping it back to ${d}$.

In our distillation framework, we use the Kullback–Leibler divergence of $\mathbf{A}_{l, a}$ between teacher and students as the distillation objective. However, matching the attention in the last layer of the students will leave the FFN in the last layer untrained as the gradient only flows backward. To this end, we propose stacking an extra layer on the students during distillation to ensure that the gradient can flow through the entire network and that all the parameters are trained during distillation. This extra layer will be discarded when the students are fine-tuned on downstream tasks.

\section{Experiments}
Three groups of experiments, including fine-tuning on natural language understanding tasks (in Sec.~\ref{sec:fine-tasks}), few-shot learning (in Sec.~\ref{sec:few-shot-tasks}), and zero-shot learning (in Sec.~\ref{sec:zero-shot-tasks}), are conducted on a variety of prevailing NLP tasks to evaluate the performance of ERNIE 3.0 Titan. All the previous state-of-the-art results for comparison come from the best public single model reported that we could find.\footnote{The previous SoTA results of ERNIE 2.0 and RoBERTa-wwm-ext on corresponding datasets are reproduced by ourselves, except for the datasets that already have released pre-trained results.}. It is essential to mention that all experimental results of ERNIE 3.0 Titan are based on the insufficiently pre-trained model so far. ERNIE 3.0 Titan is still in training, and we believe that the model will become stronger as the pre-training progresses.

\subsection{Evaluation Tasks}\label{sec:evaluation-tasks}
68 datasets belonging to 12 kinds of natural language processing tasks are used in our experiments, in which datasets marked with \textit{FC} are from FewCLUE Benchmark. Significantly, several datasets are applied to the experiments of fine-tuning / few-shot learning and zero-shot learning simultaneously in different ways. In this paper, we treat duplicate datasets used in different evaluation methods independently. The details as follows:

\begin{itemize}
\item \textbf{Sentiment Analysis (SA)}: NLPCC2014-SC~\footnote{\url{http://tcci.ccf.org.cn/conference/2014/pages/page04_dg.html}}, SE-ABSA16\_PHNS~\footnote{\url{http://alt.qcri.org/semeval2016/task5/}}, SE-ABSA16\_CAME, BDCI2019~\footnote{\url{https://www.datafountain.cn/competitions/350}},
EPRSTMT~\cite{xu2021fewclue}.
\item \textbf{Opinion Extraction (OE)}: COTE-BD~\cite{li2018cote}, COTE-MFW~\cite{li2018cote}.
\item \textbf{Natural Language Inference (NLI)}: OCNLI~\cite{xu2020clue}, CMNLI~\cite{xu2020clue}, OCNLI-FC~\cite{xu2021fewclue}.  
\item \textbf{Winograd Schema Challenge (WSC)}: CLUEWSC~\cite{xu2020clue}, CLUEWSC-FC~\cite{xu2021fewclue}.
\item \textbf{Relation Extraction (RE)}: FinRE~\cite{li2019finre}, SanWen~\cite{xu2017discourse}.
\item \textbf{Semantic Similarity (SS)}: AFQMC~\cite{xu2020clue}, LCQMC \cite{liu2018lcqmc}, PAWS-X~\cite{yang2019paws}, BQ Corpus~\cite{chen2018bq}, CSL~\cite{xu2020clue}, CSL-FC~\cite{xu2021fewclue}, BUSTM~\cite{xu2021fewclue}. 
\item \textbf{Text Classification (TC)}: TNEWS~\footnote{\url{https://github.com/aceimnorstuvwxz/toutiao-text-classfication-dataset}},
TNEWS-FC~\cite{xu2021fewclue},
IFLYTEK~\cite{co2019iflytek}, IFLYTEK-FC~\cite{xu2021fewclue},
THUCNEWS~\footnote{\url{http://thuctc.thunlp.org/}}, CNSE~\cite{liu2018matching}, CNSS~\cite{liu2018matching}, CSLDCP~\cite{xu2021fewclue}.
\item \textbf{Closed-Book Question Answering (CB-QA)}: NLPCC-DBQA~\footnote{\url{http://tcci.ccf.org.cn/conference/2016/dldoc/evagline2.pdf}}, CHIP2019, cMedQA~\cite{zhang2017cmedqa}, cMedQA2~\cite{zhang2018multi}, CKBQA~\footnote{\url{https://github.com/pkumod/CKBQA}}, WebQA~\cite{li2016webqa}.
\item \textbf{Cloze and Completion (Clz.\&Compl.)}: PD\&CFT~\cite{cui2016pdcft}, CMRC2017~\cite{cui2017dataset}, CMRC2019~\cite{cui2020sentence}, CHID~\cite{zheng2019chid}, CHID-FC~\cite{xu2021fewclue}, WPLC~\cite{ge2021chinese}.
\item \textbf{Machine Reading Comprehension (MRC)}: DRCD \cite{shao2018drcd}, DuReader \cite{he2017dureader}, Dureader$_\text{robust}$~\cite{tang2020dureaderrobust}, Dureader$_\text{checklist}$,   Dureader$_\text{yesno}$~\footnote{\url{https://aistudio.baidu.com/aistudio/competition/detail/49/?isFromLUGE=TRUE}}, C$^3$~\cite{sun2020c3}, CMRC 2018 \cite{DBLP:journals/corr/abs-1810-07366}.
\item \textbf{Legal Documents Analysis (LDA)}: CAIL2018-Task1~\cite{xiao2018cail2018}, CAIL2018-Task2~\cite{xiao2018cail2018}.
\item \textbf{Cant Understanding (CU)}: DogWhistle Insider, DogWhistle Outsider~\cite{xu-etal-2021-blow}.
\end{itemize}

\subsection{Experiments on Fine-tuning Tasks}\label{sec:fine-tasks}
The results of natural language understanding tasks are reported in Table~\ref{tab:result-understanding}. As shown in Table~\ref{tab:result-understanding}, 

\noindent\textbf{Sentiment Analysis}. Sentiment Analysis is a classification task aiming to determine whether a sentence is positive, negative, or neutral. We consider four datasets from different domains, including shopping (NLPCC2014-SC), electronics (SE-ABSA16\_PHNS, SE-ABSA16\_CAM), and financial (BDCI2019). ERNIE 3.0 Titan achieves state-of-the-art results on all four datasets.

\noindent\textbf{Opinion Extraction}. Similar to the sentiment analysis task, opinion extraction requires the model to mine the opinion of a sentence. We use two sub-datasets from Chinese Customer Review (COTE). Experiment results show that ERNIE 3.0 Titan also outperforms the current SoTA system.

\noindent\textbf{Relation Extraction}. The relation extraction task is to identify the relationship between different entities like persons and organizations. We consider FinRE and SanWen -- two relation extraction datasets for financial news and Chinese literature, respectively. ERNIE 3.0 Titan outperforms the previous SoTA model by a remarkable margin.

\noindent\textbf{Semantic Similarity}. Semantic Similarity is a classic NLP task that determines the similarity between various terms such as words, sentences, documents. In this work, we focus on sentence-level similarity tasks. We test ERNIE 3.0 Titan on several datasets in varied fields, including AFQMC, LCQMC, and BQ. Experiment results show that ERNIE 3.0 Titan outperforms the baseline models by a visible margin. 

\noindent\textbf{Text Classification}. We also evaluate ERNIE 3.0 Titan on Text classification. We consider four datasets: app descriptions (IFLYTEK) and news stories (THUCNEWS, CNSE, CNSS). Under different types of classification tasks, ERNIE 3.0 Titan can consistently achieve better accuracy.

\noindent\textbf{Closed-Book Question Answering}. Closed-Book Question Answering aims to directly answer the questions without any additional references or knowledge. We select a general QA dataset NLPCC-DBQA and two medical field datasets -- cMedQA and cMedQA2 to test the ability of ERNIE 3.0 Titan. Experiment results show that ERNIE 3.0 Titan performs better on all QA tasks. We believe knowledge-enhanced pre-training methods do bring benefits to the closed-book QA task.

\noindent\textbf{Cant Understanding}. Cant, also known as doublespeak, is an advanced language usage for humans. However, it is rather difficult for machines to understand this type of language. We test the cant understanding ability of ERNIE 3.0 Titan on DogWhistle -- a dataset based on \emph{Decrypto} game.  The model is required to select the right answer with the guidance of the corresponding cant. ERNIE 3.0 Titan gets the best result and shows its potential for understanding more difficult languages.

\noindent\textbf{Cloze and completion}. Cloze tests require the ability to understand context and vocabulary in order to identify the correct language that belongs in the deleted passages. Benefiting from the knowledge-enhanced pre-training, ERNIE 3.0 Titan achieves the best score among baselines. 

\noindent\textbf{Machine Reading Comprehension}. We comprehensively evaluate the ability of ERNIE 3.0 Titan on machine reading comprehension in different aspects, including span-predict reading comprehension (DuReader, DRCD, DuReader$_\text{checklist}$), multiple-choice reading comprehension (C3, DuReader$_\text{yesno}$), and robustness test (Dureader$_\text{robust}$). With the help of knowledge-enhanced pre-training, ERNIE 3.0 Titan surpasses the baseline models with significant enhancements on all types of tasks.

\noindent\textbf{Legal Documents Analysis}. Next, two domain-specific tasks of law are chosen to evaluate the ability of ERNIE 3.0 Titan on document analysis. These two datasets from CAIL2018 are both multi-label document classification tasks. ERNIE 3.0 Titan breaks the previous SoTA performance.

\begin{table}[]
\small
\centering
\resizebox{0.94\textwidth}{!}{
\begin{tabular}{clcclccc}
\toprule\toprule
ID                   & Task                                              & Dataset                                                  & Metric                             &                           & Previous SoTA Model     & ERNIE 3.0      & \multicolumn{1}{l}{ERNIE 3.0 Titan} \\ \midrule
\multirow{5}{*}{1}   & \multirow{5}{*}{SA}               & NLPCC2014-SC                                             & Acc.                               & \multicolumn{1}{l|}{Test} & 83.53 (SKEP)            & 86.00          & \textbf{86.32}                      \\
                     &                                                   & \multicolumn{1}{l}{SE-ABSA16\_PHNS}                      & Acc.                               & \multicolumn{1}{l|}{Test} & 82.91 (SKEP)            & 93.95          & \textbf{94.14}                      \\
                     &                                                   & SE-ABSA16\_CAME                                          & Acc.                               & \multicolumn{1}{l|}{Test} & 90.06 (SKEP)            & 96.05          & \textbf{96.47}                     \\
                     &                                                   & \multirow{2}{*}{BDCI2019}                                & \multirow{2}{*}{Acc.}              & \multicolumn{1}{l|}{Dev}  & -                       & 96.83          & \textbf{97.12}                      \\
                     &                                                   &                                                          &                                    & \multicolumn{1}{l|}{Test} & 96.26 (ERNIE 2.0)       & \textbf{97.70} & \textbf{97.70}                      \\ \midrule
\multirow{2}{*}{2}   & \multirow{2}{*}{OE}                & COTE-DP                                                  & F1                                 & \multicolumn{1}{l|}{Test} & 86.30 (SKEP)            & 92.75          &  \textbf{93.40}                              \\
                     &                                                   & COTE-MFW                                                 & F1                                 & \multicolumn{1}{l|}{Test} & 87.90 (SKEP)            & 89.90          & \textbf{90.69}                                    \\ \midrule
\multirow{4}{*}{3}   & \multirow{4}{*}{RE}              & \multirow{2}{*}{FinRE}                                   & \multirow{2}{*}{F1}                & \multicolumn{1}{l|}{Dev}  & 63.33 (ERNIE 2.0)       & 64.87          & \textbf{65.27}                      \\
                     &                                                   &                                                          &                                    & \multicolumn{1}{l|}{Test} & 60.60 (ERNIE 2.0)       & 62.88          & \textbf{63.15}                      \\
                     &                                                   & \multirow{2}{*}{SanWen}                                  & \multirow{2}{*}{F1}                & \multicolumn{1}{l|}{Dev}  & 79.92 (ERNIE 2.0)       & 81.32          & \textbf{83.07}                      \\
                     &                                                   &                                                          &                                    & \multicolumn{1}{l|}{Test} & 77.97 (ERNIE 2.0)       & 82.59          & \textbf{82.70}                      \\ \midrule
\multirow{5}{*}{4}   & \multirow{5}{*}{SS}              & AFQMC                                                    & Acc.                               & \multicolumn{1}{l|}{Dev}  & 74.92 (RoBERTa*)        & 77.02          & \textbf{77.39}                      \\
                     &                                                   & \multirow{2}{*}{LCQMC}                                   & \multirow{2}{*}{Acc.}              & \multicolumn{1}{l|}{Dev}  & -                       & 90.29          & \textbf{91.57}                      \\
                     &                                                   &                                                          &                                    & \multicolumn{1}{l|}{Test} & 89.16 (CPM-2)           & 90.38          & \textbf{90.55}                      \\
                     &                                                   & \multirow{2}{*}{BQ Corpus}                               & \multirow{2}{*}{Acc.}              & \multicolumn{1}{l|}{Dev}  & 87.11 (ZEN 2.0)         & 87.41          & \textbf{87.64}                      \\
                     &                                                   &                                                          &                                    & \multicolumn{1}{l|}{Test} & 85.99 (ZEN 2.0)         & 86.10          & \textbf{86.23}                      \\ \midrule
\multirow{7}{*}{5}   & \multirow{7}{*}{TC} & IFLYTEK                                                  & Acc.                               & \multicolumn{1}{l|}{Dev}  & 62.75 (RoBERTa*)        & 63.45          & \textbf{63.52}                      \\
                     &                                                   & \multirow{2}{*}{THUNCEWS}                                & \multirow{2}{*}{Acc.}              & \multicolumn{1}{l|}{Dev}  & 97.7  (RoBERTa*)        & 98.33          &   \textbf{98.42}                                  \\
                     &                                                   &                                                         &                                    & \multicolumn{1}{l|}{Test} & 97.6  (RoBERTa*)        & 98.66          &   \textbf{98.70}                                   \\
                     &                                                   & \multirow{2}{*}{CNSE}                                    & \multirow{2}{*}{Acc.}              & \multicolumn{1}{l|}{Dev}  & 85.64  (RoBERTa*)       & 88.94          & \textbf{90.11}                      \\
                     &                                                   &                                                          &                                    & \multicolumn{1}{l|}{Test} & 85.57  (RoBERTa*)       & 88.92          & \textbf{89.51}                      \\
                     &                                                   & \multirow{2}{*}{CNSS}                                    & \multirow{2}{*}{Acc.}              & \multicolumn{1}{l|}{Dev}  & 93.06 (ERNIE 2.0)       & 93.84          & \textbf{94.43}                      \\
                     &                                                   &                                                          &                                    & \multicolumn{1}{l|}{Test} & 92.73 (ERNIE 2.0)       & 93.76          & \textbf{94.24}                      \\ \midrule
\multirow{6}{*}{6}   & \multirow{6}{*}{CB-QA}   & \multirow{2}{*}{NLPCC-DBQA}                              & \multirow{2}{*}{MRR/F1}            & \multicolumn{1}{c|}{Dev}  & 96.04/85.69 (Zen 2.0)   & 96.71/87.57    &  \textbf{96.84/87.64}                                   \\
                     &                                                   &                                                          &                                    & \multicolumn{1}{l|}{Test} & 96.11/86.47 (Zen 2.0)   & 96.50/88.49    & \textbf{96.69/88.57}                                     \\
                     &                                                   & \multirow{2}{*}{cMedQA}                                  & \multirow{2}{*}{Acc.}              & \multicolumn{1}{l|}{Dev}  & 78.6 (BERT\_BiGRU*)     & \textbf{84.60} & \textbf{84.60}                      \\
                     &                                                   &                                                          &                                    & \multicolumn{1}{l|}{Test} & 78.2 (BERT\_BiGRU*)     & 82.65          & \textbf{83.45}                      \\
                     &                                                   & \multirow{2}{*}{cMedQA2}                                 & \multirow{2}{*}{Acc.}              & \multicolumn{1}{l|}{Dev}  & 81.3 (BERT\_BiGRU*)     & 83.48          & \textbf{83.83}                                     \\
                     &                                                   &                                                          &                                    & \multicolumn{1}{l|}{Test} & 82.2 (BERT\_BiGRU*)     & 83.68          &  \textbf{84.23}                                   \\ \midrule
\multirow{4}{*}{7}  & \multirow{4}{*}{CU}               & \multicolumn{1}{l}{\multirow{2}{*}{DogWhistle Insider}}  & \multirow{2}{*}{Acc.}              & \multicolumn{1}{c|}{Dev}  & 75.4 (ALBERT)           & \textbf{79.06} & 78.86                               \\
                     &                                                   & \multicolumn{1}{l}{}                                     &                                    & \multicolumn{1}{l|}{Test} & 76.1 (ALBERT)           & 79.22          & \textbf{79.33}                      \\
                     &                                                   & \multicolumn{1}{l}{\multirow{2}{*}{DogWhistle Outsider}} & \multirow{2}{*}{Acc.}              & \multicolumn{1}{l|}{Dev}  & 34.6 (ALBERT)           & 38.68          & \textbf{39.13}                      \\
                     &                                                   & \multicolumn{1}{l}{}                                     &                                    & \multicolumn{1}{l|}{Test} & 34.6 (ALBERT)           & 38.22          & \textbf{38.61}                      \\ \midrule
\multirow{14}{*}{8} & \multirow{14}{*}{MRC} & CRMC2019                                                 & QAC/PAC                            & \multicolumn{1}{c|}{Dev}  & 82.6/23.3 (RoBERTa*)    & 92.53/57.33    & \textbf{92.69/58.33}                \\
                     &                                                   & \multirow{2}{*}{DRCD}                                    & \multirow{2}{*}{EM/F1}             & \multicolumn{1}{c|}{Dev}  & 90.8/95.3 (MacBERT)     & 91.54/96.45    & \textbf{92.40/96.74}                \\
                     &                                                   &                                                          &                                    & \multicolumn{1}{c|}{Test} & 90.9/95.3 (MacBERT)     & 91.41/95.84    & \textbf{92.16/96.30}                \\
                     &                                                   & DuReader                                                 & EM/F1                              & \multicolumn{1}{c|}{Dev}  & 64.2/77.3 (ERNIE 2.0)   & 67.69/79.66    & \textbf{68.92/80.74}                \\
                     &                                                   & \multirow{2}{*}{DuReader$_{robust}$}                       & \multirow{2}{*}{EM/F1}             & \multicolumn{1}{c|}{Dev}  & 75.23/86.77 (ERNIE 2.0) & 77.27/88.54    & \textbf{78.12/89.12}                \\
                     &                                                   &                                                          &                                    & \multicolumn{1}{c|}{Test} & 51.20/67.96 (ERNIE 2.0) & 60.87/75.63    & \textbf{61.32/76.01}                                    \\
                     &                                                   & \multirow{2}{*}{DuReader$_{checklist}$}                    & \multirow{2}{*}{EM/F1}             & \multicolumn{1}{c|}{Dev}  & 55.66/64.12 (ERNIE 2.0) & 61.33/70.59    & \textbf{61.59/71.19}                \\
                     &                                                   &                                                          &                                    & \multicolumn{1}{c|}{Test} & 59.11/48.79 (ERNIE 2.0) & 64.87/53.82    &  \textbf{64.98/54.02}                                   \\
                     &                                                   & \multirow{2}{*}{DuReader$_{yesno}$}                        & \multirow{2}{*}{Acc.}              & \multicolumn{1}{c|}{Dev}  & 88.69 (ERNIE 2.0)       & 89.95          &  \textbf{90.04}                                   \\
                     &                                                   &                                                          &                                    & \multicolumn{1}{c|}{Test} & 88.82 (ERNIE 2.0)       & 89.64          &  \textbf{90.31}                                   \\
                     &                                                   & \multirow{2}{*}{C3}                                      & \multirow{2}{*}{Acc.}              & \multicolumn{1}{c|}{Dev}  & -                       & 87.63          & \textbf{88.34}                      \\
                     &                                                   &                                                          &                                    & \multicolumn{1}{c|}{Test} & 86.1 (CPM-2)            & 86.69          & \textbf{87.59}                      \\
                     &                                                   & CHID                                                     & Acc.                               & \multicolumn{1}{c|}{Dev}  & 85.81 (RoBERTa*)        & 91.67          & \textbf{92.58}                      \\ \midrule
\multirow{4}{*}{9}  & \multirow{4}{*}{LDA}          & \multirow{2}{*}{CAIL2018 Task1}                          & \multirow{2}{*}{F1-macro/F1-micro} & \multicolumn{1}{c|}{Dev}  & 83.85/91.50 (ERNIE 2.0) & 88.64/93.11    & \textbf{88.69/93.18}                                    \\
                     &                                                   &                                                          &                                    & \multicolumn{1}{c|}{Test} & 80.40/89.94 (ERNIE 2.0) & 86.83/91.82    &  \textbf{86.88/91.87}                                    \\
                     &                                                   & \multirow{2}{*}{CAIL2018 Task2}                          & \multirow{2}{*}{F1-macro/F1-micro} & \multicolumn{1}{c|}{Dev}  & 78.58/89.46 (ERNIE 2.0) & 82.62/90.93    &  \textbf{82.66/90.95}                                    \\
                     &                                                   &                                                          &                                    & \multicolumn{1}{c|}{Test} & 75.35/86.97 (ERNIE 2.0) & 81.10/88.52    &  \textbf{81.12/88.61}                 \\ \bottomrule\bottomrule
\end{tabular}
}
\\[5pt]
\caption{Results on Natural Language Understanding Tasks. We compare ERNIE 3.0 Titan with ERNIE 3.0 and 10 previous SoTA baselines including CPM-2~\cite{zhang2021cpm}, ERNIE 2.0~\cite{sun2020ernie}, SKEP~\cite{tian2020skep}, RoBERTa-wwm-ext-large~\cite{cui2019pre} (marked as RoBERTa*), ALBERT~\cite{lan2019albert}, MacBERT~\cite{cui2020revisiting}, Zen 2.0~\cite{song2021zen} and crossed BERT siamese BiGRU~\cite{cui2020chinese} (marked as BERT\_BiGRU*).}\label{tab:result-understanding}
\end{table}

\subsection{Experiments on Few-shot Learning}\label{sec:few-shot-tasks}
\subsubsection{Settings}
In this section, we evaluate the few-shot performance of ERNIE 3.0 Titan on the FewCLUE benchmark. FewCLUE is a comprehensive few-shot evaluation benchmark in Chinese, including various of tasks. Flowing Section \ref{sec:evaluation-tasks}, we categorize these task into six different types including text classification, natural language understanding etc. Each task provides five training/evaluation splits and a corresponding union set (called train$\_$all and dev$\_$all).  The number of the few-shot training samples is related to the number of the label categories. In other words, tasks with more classes have more training samples. We choose mT5-XXL, Yuan 13B-PLM, and ERNIE 3.0 as our baselines.  For mT5-XXL, the results are produced based on the open-source codes with their default tuning method, and for Yuan 13B-PLM, we directly use the published results.

We tested three approaches to train the few-shot learners based on ERNIE 3.0 Titan. For different types of tasks, we use the corresponding task-specific training method:
\begin{itemize}
\item For text classification, sentiment analysis, and semantic similarity tasks, we use traditional fine-tuning methods.
\item For reading comprehension (CHID-FC) and winograd schema challenge task, we utilize pattern exploiting training~\cite{schick2020exploiting} to reformulate such tasks cloze-style questions and perform gradient-based tuning.
\item For the natural language inference task, we exploit NSP-based prompt training~\cite{sun2021nsp}. Different labels are regarded as prompts to concatenate the two sentences, and the model is trained to select the label that makes the concatenated sentence the most coherent. The classification network of the sentence distance task (Section \ref{sec:Structure-aware pretrain-task}) in the pre-training phase can be seen as a good initialization to train the coherent ranker.
\end{itemize}

We use the union set to train the few-shot learner and report the results on the evaluation sets (dev$\_$all) and public test sets to conduct the experiments. 

\begin{table}[]
\small
\centering
\resizebox{0.8\textwidth}{!}{
\begin{tabular}{lccccccc}
\toprule\toprule
\textbf{Task Type}            & \textbf{Dataset}            & \textbf{Metric}       & \textbf{}                 & \textbf{mT5-XXL} & \textbf{Yuan 1.0} & \textbf{ERNIE 3.0} & \textbf{ERNIE 3.0 Titan} \\ \midrule
\multirow{6}{*}{\textbf{TC}}  & \multirow{2}{*}{TNEWS-FC}   & \multirow{2}{*}{Acc.} & \multicolumn{1}{c|}{Dev}  & 62.80            & 56.74             & 63.30              & \textbf{65.21}           \\
                              &                             &                       & \multicolumn{1}{c|}{Test} & 65.50            & -                 & 64.83              & \textbf{67.26}           \\
                              & \multirow{2}{*}{IFLYTEK-FC} & \multirow{2}{*}{Acc.} & \multicolumn{1}{c|}{Dev}  & 53.56            & 57.78             & 58.41              & \textbf{59.29}           \\
                              &                             &                       & \multicolumn{1}{c|}{Test} & 54.87            & -                 & 57.75              & \textbf{59.92}           \\
                              & \multirow{2}{*}{CSLDCP}     & \multirow{2}{*}{Acc.} & \multicolumn{1}{c|}{Dev}  & 65.25            & 66.44             & 68.76              & \textbf{70.21}           \\
                              &                             &                       & \multicolumn{1}{c|}{Test} & 65.83            & -                 & 68.05              & \textbf{69.84}           \\ \midrule
\multirow{4}{*}{\textbf{SS}}  & \multirow{2}{*}{CSL-FC}     & \multirow{2}{*}{Acc.} & \multicolumn{1}{c|}{Dev}  & 74.38            & 83.75             & 79.38              & \textbf{86.25}           \\
                              &                             &                       & \multicolumn{1}{c|}{Test} & 75.02            & -                 & 75.48              & \textbf{76.22}           \\
                              & \multirow{2}{*}{BUSTM}      & \multirow{2}{*}{Acc.} & \multicolumn{1}{c|}{Dev}  & 70.63            & 83.75             & 83.75              & \textbf{84.38}           \\
                              &                             &                       & \multicolumn{1}{c|}{Test} & 69.46            & -                 & 76.69              & \textbf{77.09}           \\ \midrule
\multirow{2}{*}{\textbf{SA}}  & \multirow{2}{*}{EPRSTMT}    & \multirow{2}{*}{Acc.} & \multicolumn{1}{c|}{Dev}  & 84.84            & 90.63             & \textbf{91.25}     & \textbf{91.25}           \\
                              &                             &                       & \multicolumn{1}{c|}{Test} & 86.48            & -                 & 88.36              & \textbf{90.00}           \\ \midrule
\multirow{2}{*}{\textbf{NLI}} & \multirow{2}{*}{OCNLI-FC}   & \multirow{2}{*}{Acc.} & \multicolumn{1}{c|}{Dev}  & 41.50            & 58.13             & 58.13              & \textbf{65.00}           \\
                              &                             &                       & \multicolumn{1}{c|}{Test} & 41.51            & -                 & 58.69              & \textbf{59.77}           \\ \midrule
\multirow{2}{*}{\textbf{WSC}} & \multirow{2}{*}{CLUEWSC-FC} & \multirow{2}{*}{Acc.} & \multicolumn{1}{c|}{Dev}  & 76.47            & 82.39             & 84.28              & \textbf{90.57}           \\
                              &                             &                       & \multicolumn{1}{c|}{Test} & 71.52            & -                 & 82.48              & \textbf{85.66}           \\ \midrule
\multirow{2}{*}{\textbf{MRC}} & \multirow{2}{*}{CHID-FC}    & \multirow{2}{*}{Acc.} & \multicolumn{1}{c|}{Dev}  & 46.63            & 76.24             & 80.69              & \textbf{84.16}           \\
                              &                             &                       & \multicolumn{1}{c|}{Test} & 46.02            & -                 & 83.72              & \textbf{85.21}           \\ \bottomrule\bottomrule
\end{tabular}
}
\\[5pt]
\caption{Results on Few-shot learning tasks.}
\label{tab:few-shot-result}
\end{table}

\subsubsection{Results} 
The main results of the few-shot learning tasks are illustrated in Table \ref{tab:few-shot-result}. ERNIE 3.0 Titan consistently outperforms baseline models, including ERNIE 3.0. Under vanilla fine-tuning methods, ERNIE 3.0 Titan still can surpass Yuan 1.0 4.58\% average points on the text classification task, 1.56\% on sentence similarity, and 0.62\% on semantic similarity tasks. The task of reading comprehension and winograd schema challenge can be naturally reformulated to cloze-style tasks. Under prompt-based pattern exploiting, ERNIE 3.0 Titan outperforms Yuan 1.0 by a remarkable margin: 8.18\% absolute improvement on CLUEWSC-FC and 7.92\% on CHID-FC dataset. For natural language inference task, ERNIE 3.0 Titan also achieves a significant improvement of 6.87 points.

\begin{table}[]
\centering
\resizebox{\textwidth}{!}{%
\begin{tabular}{@{}lcccccccccc@{}}
\toprule \toprule
\textbf{Task Type}                     & \textbf{Dataset} & \textbf{Metric}                  & \textbf{CPM-1}   & \textbf{PanGu-$\alpha$-2.6B} & \textbf{PanGu-$\alpha$-13B} & \textbf{Yuan 1.0-13B} & \textbf{Yuan 1.0-245B} & \textbf{GPT-3} & \textbf{ERNIE 3.0} & \textbf{ERNIE 3.0 Titan}    \\ \midrule
\multirow{5}{*}{\textbf{TC}}           & TNEWS            & \multicolumn{1}{c|}{Acc.}        & 65.44            & 60.95                        & 60.26                       & -                     & -                      & -              & 68.40              & \textbf{72.60}              \\
                                       & TNEWS-FC         & \multicolumn{1}{c|}{Acc.}        & -                & -                            & -                           & 57.47                 & 57.19                  & -              & -                  & \textbf{57.83}              \\
                                       & IFLYTEK          & \multicolumn{1}{c|}{Acc.}        & 68.91            & 74.26                        & 73.80                       & -                     & -                      & -              & 75.34              & \textbf{79.84}              \\
                                       & IFLYTEK-FC       & \multicolumn{1}{c|}{Acc.}        & -                & -                            & -                           & 38.82                 & -                      & -              & -                  & \textbf{40.42}              \\
                                       & CSLDCP           & \multicolumn{1}{c|}{Acc.}        & -                & -                            & -                           & 47.53                 & 48.02                  & -              & -                  & \textbf{50.97}              \\ \midrule
\textbf{SA}                            & EPRSTMT          & \multicolumn{1}{c|}{Acc.}        & -                & -                            & -                           & 88.13                 & 86.88                  & -              & -                  & \textbf{88.75}              \\ \midrule
\multirow{4}{*}{\textbf{SS}}           & AFQMC            & \multicolumn{1}{c|}{Acc.}        & 66.34            & 59.29                        & 65.76                       & -                     & -                      & -              & \textbf{68.99}     & \textbf{68.99}              \\
                                       & CSL              & \multicolumn{1}{c|}{Acc.}        & 52.30            & 50.50                        & 49.30                       & -                     & -                      & -              & 55.63              & \textbf{55.80}              \\
                                       & CSL-FC           & \multicolumn{1}{c|}{Acc.}        & -                & -                            & -                           & 50.00                 & -                      & -              & -                  & \textbf{56.25}              \\
                                       & BUSTM            & \multicolumn{1}{c|}{Acc.}        & -                & -                            & -                           & 59.38                 & -                      & -              & -                  & \textbf{64.38}              \\ \midrule
\multirow{3}{*}{\textbf{NLI}}          & OCNLI            & \multicolumn{1}{c|}{Acc.}        & 44.20            & 42.61                        & 41.53                       & -                     & -                      & -              & 44.31              & \textbf{44.61}              \\
                                       & OCNLI-FC         & \multicolumn{1}{c|}{Acc.}        & -                & -                            & -                           & 48.13                 & -                      & -              & -                  & \textbf{53.75}              \\
                                       & CMNLI            & \multicolumn{1}{c|}{Acc.}        & 49.10            & 50.20                        & 48.44                       & -                     & -                      & -              & 49.41              & \textbf{51.70}              \\ \midrule
\multirow{2}{*}{\textbf{WSC}}          & CLUEWSC          & \multicolumn{1}{c|}{Acc.}        & 73.68            & 73.36                        & 75.00                       & -                     & -                      & -              & 78.38              & \textbf{81.08}              \\
                                       & CLUEWSC-FC       & \multicolumn{1}{c|}{Acc.}        & -                & -                            & -                           & 38.99                 & -                      & -              & -                  & \textbf{53.46}              \\ \midrule
\multirow{7}{*}{\textbf{Clz.\&Compl.}} & CHID             & \multicolumn{1}{c|}{Acc.}        & 68.62            & 68.73                        & 70.64                       & -                     & -                      & -              & 77.78              & \textbf{86.21}              \\
                                       & CHID-FC          & \multicolumn{1}{c|}{Acc.}        & -                & -                            & -                           & 86.14                 & -                      & -              & -                  & \textbf{87.13}              \\
                                       & PD               & \multicolumn{1}{c|}{Acc.}        & 35.73            & 38.47                        & 43.84                       & -                     & -                      & -              & 66.07              & \textbf{67.06}              \\
                                       & CFT              & \multicolumn{1}{c|}{Acc.}        & 38.99            & 42.39                        & 46.60                       & -                     & -                      & -              & 49.30              & \textbf{66.14}              \\
                                       & CMRC2017         & \multicolumn{1}{c|}{Acc.}        & 24.60            & 37.83                        & 38.90                       & -                     & -                      & -              & 56.66              & \textbf{74.63}              \\
                                       & CMRC2019         & \multicolumn{1}{c|}{Acc.}        & 47.69            & 61.93                        & 68.19                       & -                     & -                      & -              & \textbf{75.00}     & \textbf{75.00}              \\
                                       & WPLC             & \multicolumn{1}{c|}{PPL}         & -                & 48.98                        & 45.85                       & -                     & -                      & -              & 17.03              & \textbf{16.50}              \\ \midrule
\multirow{4}{*}{\textbf{MRC}}          & C3               & \multicolumn{1}{c|}{Acc.}        & 49.81            & 53.42                        & 54.47                       & -                     & -                      & -              & 52.62              & \textbf{54.85}              \\
                                       & CMRC2018         & \multicolumn{1}{c|}{Avg.(EM/F1)} & 5.36(0.59/10.12) & 8.93(1.21/16.65)             & 10.37(1.46/19.28)           & -                     & 27.38(5.58/49.17)      & -              & 16.61(7.61/25.61)  & \textbf{30.41}(16.62/44.20) \\
                                       & DRCD             & \multicolumn{1}{c|}{Avg.(EM/F1)} & 2.31(0.00/4.62)  & 5.40(0.80/9.99)              & 5.61(0.66/10.55)            & -                     & -                      & -              & 18.44(10.58/26.29) & \textbf{29.46}(21.08/37.83) \\
                                       & DuReader         & \multicolumn{1}{c|}{ROUGE-1}     & 16.63            & 21.07                        & 24.46                       & -                     & -                      & -              & 29.79              & \textbf{32.13}              \\ \midrule
\multirow{3}{*}{\textbf{CB-QA}}        & WebQA            & \multicolumn{1}{c|}{Avg.(EM/F1)} & 9.30(6.00/12.59) & 9.07(4.43/13.71)             & 9.80(5.13/14.47)            & -                     & 40.47(30.57/50.36)     & -              & 30.74(22.53/38.95) & \textbf{45.27}(37.97/52.57) \\
                                       & CKBQA            & \multicolumn{1}{c|}{Acc.}        & 13.40            & 14.61                        & 14.21                       & -                     & -                      & -              & 20.64              & \textbf{24.12}              \\
                                       & CKBQA-sub        & \multicolumn{1}{c|}{Acc.}        & 13.52            & 14.04                        & 13.92                       & -                     & -                      & 14.76          & 18.29              & \textbf{22.84}              \\ \bottomrule \bottomrule
\end{tabular}%
}
\\[5pt]
\caption{Results on zero-shot learning tasks. We reported the results on the dev-all set.}
\label{tab:zero-shot-result}
\end{table}

\subsection{Experiments on Zero-shot Learning}\label{sec:zero-shot-tasks} 
%We have demonstrated that ERNIE 3.0 Titan is superior to previous SoTA methods on both NLU and NLG tasks following the pretraining-then-finetuning paradigm. 
This section conducts various types of tasks with the zero-shot setting where a model is applied without parameter updates. ERNIE 3.0 Titan achieves strong performance compared to recently proposed large-scale Chinese language models such as CPM-1 (2.6B), PanGu-$\alpha$, Yuan 1.0 on all downstream tasks. On the CKBQA-sub dataset, which requires strong knowledge reasoning ability, ERNIE 3.0 Titan surpassed GPT-3 by over 8 point percent with respect to accuracy. In our case study (Sec.~\ref{sec: zero-shot-case-study}), we demonstrate the ability of ERNIE 3.0 Titan to generate controllable and credible results. Quantitatively, we evaluated ERNIE 3.0 with baselines on our manually collected 467 cases across 13 different tasks and showed that it could generate more coherent, natural, and accurate responses. 

\subsubsection{Evaluation Methods} 

This section unified five probability forms of the scoring function for tasks with a limited label set, such as text classification, sentiment analysis, and cloze-style tasks. Based on the unified ERNIE 3.0 framework, the implementation of these five scoring functions can be task-specific. The ablation study in Sec.~\ref{sec: analysis-scoring-function} shows the effect of different scoring functions where some can obtain stable performance gain over others on a certain task type.

\begin{table}[]
\centering
\resizebox{\textwidth}{!}{%
\begin{tabular}{@{}lcc@{}}
\toprule \toprule
\textbf{Probability Form}                       & \textbf{Implementation}                                                                                                                                                                                                                                                                    & \textbf{Example}                                                                                    \\ \midrule
\multirow{2}{*}{$P(x,y)$}              & \multirow{2}{*}{$\arg\max_{i}\frac{\sum _{j=0}^{_{l_{i}}}\log P\left( f_{\text{fill}}\left( x^{'} ,y_{i}\right)^{j} |f_{\text{fill}}\left( x^{'} ,y_{i}\right)^{< j}\right)}{l_{i} \ }$}                                                                                          & $x^{'} =\underline{\text{hypothesis?}y_i\text{, premise.}}$                                \\
                                       &                                                                                                                                                                                                                                                                                   & $y_{i} \in \left\{\text{Yes,No,Maybe}\right\}$                                             \\ \midrule
\multirow{2}{*}{$P(y|x)$}              & \multirow{2}{*}{$\arg\max_{i}\frac{\sum _{j=0}^{_{|y_{i} |}}\log P\left( y{_{i}^{j}} |f_{\text{fill}}\left( x^{'} ,y_{i}^{< j}\right)\right)}{|y_{i} |}$}                                                                                                                         & $x^{'} =\text{News:} x.\text{ This news is about }\underline{y_i}\text{.}$                 \\
                                       &                                                                                                                                                                                                                                                                                   & $y_{i} \ \in \left\{\text{culture, sports, tech., etc.}\right\}$                           \\ \midrule
\multirow{2}{*}{$P(x|y)$}              & \multirow{2}{*}{$\arg\max_{i}\sum _{j=0}^{_{|x^{'} |}}\log P\left( x^{'^{j}} |f_{\text{fill}}\left( x^{'^{< j}} ,y_{i}\right)\right)$}                                                                                                                                            & $x^{'} =\text{This news is about }y_i\text{. }\text{News:} \underline{x}.$                 \\
                                       &                                                                                                                                                                                                                                                                                   & $y_{i} \ \in \left\{\text{culture, sports, tech., etc.}\right\}$                           \\ \midrule
\multirow{3}{*}{$\frac{P(y|x)}{P(y)}$} & \multirow{3}{*}{$\arg\max_{i}\frac{\sum _{j=0}^{_{|y_{i} |}}\log P\left( y{_{i}^{j}} |f_{\text{fill}}\left( x^{'} ,y_{i}^{< j}\right)\right) \ -\ \sum _{j=0}^{_{|y_{i} |}}\log P\left( y{_{i}^{j}} |f_{\text{fill}}\left( \emptyset ^{'} ,y_{i}^{< j}\right)\right)}{|y_{i} |}$} & $x^{'} =\text{News:} x.\text{ This news is about }\underline{y_i}\text{.}$                 \\
                                       &                                                                                                                                                                                                                                                                                   & $\emptyset^{'} =\text{News:} \emptyset.\text{ This news is about }\underline{y_i}\text{.}$ \\
                                       &                                                                                                                                                                                                                                                                                   & $y_{i} \ \in \left\{\text{culture, sports, tech., etc.}\right\}$                           \\ \midrule
\multirow{2}{*}{$P(\text{True}|x, y)$} & \multirow{2}{*}{$\arg\max_{i} P\left(\text{True} |f_{\text{fill}}\left( x^{'} ,y_{i}\right)\right)$}                                                                                                                                                                              & $x^{'} =\text{\underline{\texttt{[CLS]}} hypothesis? \texttt{[SEP]} }y_i\text{, premise.}$ \\
                                       &                                                                                                                                                                                                                                                                                   & $y_{i} \in \left\{\text{Yes,No,Maybe}\right\}$                                             \\ \bottomrule  \bottomrule
\end{tabular}%
}
\\[5pt]
\caption{Notations and scoring functions for zero-shot learning.}
\label{tab:zero-shot-eval}
\end{table}

In Table.~\ref{tab:zero-shot-eval}, five forms of the scoring function are shown where $f_{\text{prompt}}(\cdot)$ is the function to prompt the input $x$ to $x^{'}$, $f_{\text{fill}(\cdot)}$ is the function to fill the $i_\text{th}$ label into prompted text $x^{'}$, $y_i \in \mathcal{Y}$, $\mathcal{Y}$ is the label set and $|y_i|, l_i$ denote the tokens length for label $y_i$ and the filled prompt respectively. We will introduce these scoring functions as follows:
\begin{itemize}
    \item \textbf{$P(x, y)$}. The implementation of the joint probability $P(x, y)$ is equivalent to the perplexity of the prompted text, meaning that the prompted text with the lowest perplexity score will be predicted as the correct answer. 
    \item \textbf{$P(y|x)$}. Given the input text $x$, we will choose the highest probability answer among all possible labels. To ease the effect of the label length bias, length normalization is commonly used. This method is also called the average log-likelihood used in GPT-3~\cite{gpt-3}.
    \item \textbf{$P(x|y)$} is the reverse version of $P(y | x)$. The above two forms explicitly include the label's probability, which ignores the effect of the label bias. Since the label is imbalanced distributed in the pre-training corpus, the model tends to assign a higher probability to labels common in the corpus. By conditioning on $y$, we assume the impact of the label bias will be flattened over the input tokens.
    \item \textbf{$\frac{P(y|x)}{P(y)}$} is proportional to the pointwise mutual information (PMI) of $x, y$. PMI has been used for finding collocations and associations between words. Compared to $P(y|x)$, we can think of $\frac{P(y|x)}{P(y)}$ as a way to eliminate the effect of label bias, since common labels with high probability will decrease the final score through division. In practice, it is better to restrain the $P(y)$ in the target domain by prompting the text using a domain-specific prompt but taking as input an empty input which is used in~\cite{AriHoltzman2021SurfaceFC}.
    \item $P(\text{True}|x, y)$ formalizes each possible answer with the input text as a binary classification task. In this way, the next sentence prediction (NSP) task~\footnote{In ERNIE framework, we use the sentence distance prediction task which is an extension of the traditional next sentence prediction (NSP) task (introduced in Sec.~\ref{sec:Structure-aware pretrain-task}).} can be utilized, which has been pre-trained on the hidden state of \texttt{[CLS]}. Intuitively, NSP task estimates the affinity score of two sentences. Thus, we can prompt the text into two sentences where one is filled with different labels.
\end{itemize}

For generative tasks such as machine reading comprehension, ERNIE 3.0 used a restrained beam search with a beam width of 8 for extractive MRC to ensure the generation is a span that occurred in the context. Though effective, the beam search is time-consuming for ERNIE 3.0 Titan. Since ERNIE 3.0 Titan is assumed to be more powerful, we used the top-1 sampling strategy for all generation tasks.

\subsubsection{Results}

\begin{table}[]
\centering
\resizebox{\textwidth}{!}{%
\begin{tabular}{@{}lcccc@{}}
\toprule \toprule
\textbf{Task Type}                     & \textbf{Dataset} & \textbf{Prompting Function}                                                                                       & \textbf{Label Set}                                            & \textbf{Eval. Method} \\ \midrule
\multirow{5}{*}{\textbf{TC}}           & TNEWS            & This is the article about $y_i$: $x$                                                                              & $y_i \in \{\text{story, culture, entertainment, etc.}\}$      & P(x|y)                \\
                                       & TNEWS-FC         & News:$x$. This news is about $y_i$.                                                                               & $y_i \in \{\text{story, culture, entertainment, etc.}\}$      & P(y|x)/P(y)           \\
                                       & IFLYTEK          & This is the application about $y_i$: $x$                                                                          & $y_i \in \{\text{map, free wifi, instant messaging, etc.}\}$  & P(x|y)                \\
                                       & IFLYTEK-FC       & This is the application about $y_i$: $x$                                                                          & $y_i \in \{\text{map, free wifi, instant messaging, etc.}\}$  & P(y|x)/P(y)           \\
                                       & CSLDCP           & The subject to which the paper belongs is $y_i$. Paper: $x$                                                       & $y_i \in \{\text{materialogy, crop science, oralogy, etc.}\}$ & P(y|x)/P(y)           \\ \midrule
\textbf{SA}                            & EPRSTMT          & $y_i$, $x$                                                                                                        & $y_i \in \{\text{Great, Awful}\}$                             & P(x|y)                \\ \midrule
\multirow{4}{*}{\textbf{SS}}           & AFQMC            & $x_\texttt{text-a}$. $x_\texttt{text-b}$. The semantic meaning of the above two sentences are $y_i$.              & $y_i \in \{\text{different, the same}\}$                      & P(y|x)                \\
                                       & CSL              & \texttt{[web0] [web1] $\cdots$ [web63]} Abstract: $x_\texttt{abstract}$ The keywords $y_i$: $x_\texttt{keywords}$ & $y_i \in \{\text{aren't, are}\}$                              & P(x|y)                \\
                                       & CSL-FC           & \texttt{[web0] [web1] $\cdots$ [web63]} Abstract: $x_\texttt{abstract}$ The keywords $y_i$: $x_\texttt{keywords}$ & $y_i \in \{\text{aren't, are}\}$                              & P(x|y)                \\
                                       & BUSTM            & \texttt{[CLS]} $x_\texttt{text-a}$? \texttt{[SEP]}$y_i$, $x_\texttt{text-b}$                                      & $y_i \in \{\text{it is not, it is}\}$                         & P(True|x, y)          \\ \midrule
\multirow{3}{*}{\textbf{NLI}}          & OCNLI            & \texttt{[CLS]} $x_\texttt{hypothesis}$? \texttt{[SEP]}$y_i$, $x_\texttt{premise}$                                 & $y_i \in \{\text{Yes, Maybe, No}\}$                           & P(True|x, y)          \\
                                       & OCNLI-FC         & \texttt{[CLS]} $x_\texttt{hypothesis}$? \texttt{[SEP]}$y_i$, $x_\texttt{premise}$                                 & $y_i \in \{\text{Yes, Maybe, No}\}$                           & P(True|x, y)          \\
                                       & CMNLI            & \texttt{[novel0] [novel1] $\cdots$ [novel63]} $x_\texttt{hypothesis}$? $y_i$, $x_\texttt{premise}$                & $y_i \in \{\text{Yes, Maybe, No}\}$                           & P(x|y)                \\ \midrule
\multirow{2}{*}{\textbf{WSC}}          & CLUEWSC          & $x_\texttt{left}$ $y_i$ $x_\texttt{right}$                                                                        & $y_i \in \mathcal{Y}$                                       & P(y|x)                \\
                                       & CLUEWSC-FC       & $x_\texttt{left}$ ($y_i$ $x_\texttt{span-text}$) $x_\texttt{right}$                                               & $y_i \in \{\text{i.e., not}\}$                                & P(x|y)                \\ \midrule
\multirow{7}{*}{\textbf{Clz.\&Compl.}} & CHID             & \texttt{[web0] [web1] $\cdots$ [web63]} $x_\texttt{left}$ $y_i$ $x_\texttt{right}$                                & $y_i \in \mathcal{Y}$                                       & P(x|y)                \\
                                       & CHID-FC          & $x_\texttt{left}$ $y_i$ $x_\texttt{right}$                                                                        & $y_i \in \mathcal{Y}$                                       & P(y|x)                \\
                                       & PD               & $x_\texttt{left}$ $y_i$ $x_\texttt{right}$                                                                        & $y_i \in x$                                                   & Restrained-Gen        \\
                                       & CFT              & $x_\texttt{left}$ $y_i$ $x_\texttt{right}$                                                                        & $y_i \in x$                                                   & Restrained-Gen        \\
                                       & CMRC2017         & $x_\texttt{left}$ $y_i$ $x_\texttt{right}$                                                                        & $y_i \in x$                                                   & Restrained-Gen        \\
                                       & CMRC2019         & $x_\texttt{left}$ $y_i$ $x_\texttt{right}$                                                                        & $y_i \in \mathcal{Y}$                                       & P(x|y)                \\
                                       & WPLC             & \texttt{[novel0] [novel1] ... [novel64]} $x$                                                                      & -                                                             & PPL.                  \\ \midrule
\multirow{4}{*}{\textbf{MRC}}          & C3               & Question: $x_\texttt{question}$ Answer: $y_i$. Reading the article: $x_\texttt{article}$                          & $y_i \in \mathcal{Y}$                                       & P(y|x)/P(y)           \\
                                       & CMRC2018         & Reading the article: $x_\texttt{article}$ Question: $x_\texttt{question}$ Answer:                          & $y_i \in x_\texttt{article}$                                  & Top1-Gen              \\
                                       & DRCD             & Reading the article: $x_\texttt{article}$ Question: $x_\texttt{question}$ Answer:                          & $y_i \in x_\texttt{article}$                                  & Top1-Gen              \\
                                       & DuReader         & Reading the article: $x_\texttt{article}$ Question: $x_\texttt{question}$ Answer:                          & -                                                             & Top1-Gen              \\ \midrule
\multirow{3}{*}{\textbf{CB-QA}}        & WebQA            & Question: $x$ Answer:                                                                                      & -                                                             & Top1-Gen              \\
                                       & CKBQA            & Question: $x$ Answer:                                                                                      & -                                                             & Top1-Gen              \\
                                       & CKBQA-sub        & Question: $x$ Answer:                                                                                      & -                                                             & Top1-Gen              \\ \bottomrule \bottomrule
\end{tabular}%
}
\\[5pt]
\caption{Prompts and evaluation methods of ERNIE 3.0 Titan used in zero-shot learning tasks.}
\label{tab:prompt-zero-shot}
\end{table}

In Table.~\ref{tab:prompt-zero-shot}, we summarized the prompting functions, label verbalizers and evaluation methods we used for each task.
The detailed results will be introduced as follows:

\noindent\textbf{Text Classification}. For the TNEWS and IFLYTEK datasets, we randomly sample three candidates as negative labels for each sample. This sampling strategy is aligned with CPM-1's, PanGu-$\alpha$'s, and ERNIE 3.0's to make fair comparisons. While for TNEWS-FC, IFLYTEK-FC, CSLDCP, the negative sampling strategy is discarded in order to compare with Yuan 1.0. ERNIE 3.0 outperforms competitive baselines on these tasks.

\noindent\textbf{Sentiment analysis}. On the EPRSTMT dataset, ERNIE 3.0 Titan achieves 88.75\% w.r.t. accuracy on the zero-shot setting, meaning sentiment analysis is a simple task for large-scale pre-trained models.

\noindent\textbf{Semantic Similarity}. We consider AFQMC, CSL, CSL-FC, and BUSTM datasets. ERNIE 3.0 Titan outperforms baselines at a large margin. However, compared to ERNIE 3.0, the performance gain on AFQMC and CSL is marginal, and the minor improvement on CSL comes from the soft prompt tokens $\texttt{[webN]}$. It means that the soft prompt tokens appended before the original text help the model utilize the knowledge in a specific domain. On the other hand, it is hard for models to learn the semantic similarity based only on a language modeling loss. While on BUSTM, ERNIE 3.0 Titan using the NSP task as the scoring function surpassed Yuan 1.0 by 5\% point. We assume the inductive bias of the NSP task helps the model learn semantic similarity.

\noindent\textbf{Natural Language Inference}. ERNIE 3.0 Titan is evaluated on three NLI datasets, namely OCNLI, OCNLI-FC, and CMNLI, and achieves the best performance. We used the $P(\text{True}|x, y)$ scoring function for OCNLI-FC since we found that the NSP task shows some capability in modeling the semantic similarity on the BUSTM dataset. We tested different soft prompt tokens for CMNLI including \texttt{[webN]}, \texttt{[qaN]} and \texttt{[novelN]}. Soft prompt tokens \texttt{[novelN]} achieves the highest score (51.70) compared to 49.87 with \texttt{[webN]} and 49.4 with \texttt{[qaN]}. And, there is still a large room for improvement for pre-trained models on zero-shot NLI tasks. 

\noindent\textbf{Winograd Schema Challenge}: We formalize the CLUEWSC dataset as a multi-choice completion task where a pronoun is replaced with each candidate to calculate the scores. Since there are no candidates set for the CLUEWSC-FC dataset, we come up with a superior prompt by appending a complement after the target pronoun, and the model is required to judge the correctness of the complement. ERNIE 3.0 Titan surpassed Yuan 1.0 a lot on the CLUEWSC-FC dataset and achieved superior performance on CLUEWSC attributing to the power of scale.

\noindent\textbf{Cloze and completion}. We split a sample containing multiple blanks as multiple sentences to predict independently on the CHID dataset. The Hungarian algorithm~\cite{kuhn1955hungarian} is used to ensure that two blanks in one sample have unique predictions. ERNIE 3.0 Titan achieves the best score among baselines, and the Hungarian algorithm contributes a lot improving from 77.32 to 86.21. For Chinese Word Prediction with Long Context (Chinese WPLC), a sample consists of a masked text and a correct word. Following PanGu-$\alpha$, we replace the mask token with the correct word and calculate the perplexity score of a whole sentence. ERNIE 3.0 Titan achieves a much lower perplexity score (16.50) with the help of soft prompt tokens \texttt{[novelN]}. On the CMRC2019 dataset, we randomly sample three negative candidates for each blank from the original candidates, then beam search is applied to calculate the optimal path for a sample. We also formalize the PD, CFT, and CMRC2017 as multi-choice tasks where multiple choices are the words appearing in the given text. For efficiency, restricted generation~\cite{sun2021ernie} is used for these three datasets. ERNIE 3.0 Titan surpassed the baselines with a large margin. 

\noindent\textbf{Machine Reading Comprehension}. We consider four MRC datasets. Due to the power of ERNIE 3.0 Titan, we simply utilized the top-1 sampling strategy to generate answers. The maximum generated length of completion is limited by a pre-defined number based on 95\% percentile point of answers' length on the dataset. The performance of ERNIE 3.0 Titan is superior, outperforming Yuan 1.0 with a comparable number of parameters by 3.03\% point on average for the CMRC2018 dataset.

\noindent\textbf{Closed-book Question Answering}. We evaluated ERNIE 3.0 on two Closed-book Question Answering datasets which require the model to generate answers using its inherent knowledge learned during pre-training. WebQA is a large-scale real-world QA dataset from Baidu Zhidao. CKBQA is a knowledge-based question answering task. We only provide ERNIE 3.0 Titan with the question without additional evidence. In addition, we evaluated GPT-3 on a subset of the CKBQA dataset where questions requiring the background knowledge of China are filtered out, and then questions are manually translated into English. The engine for GPT-3 is Davinci, and the object is text completion. ERNIE 3.0 Titan significantly outperforms baselines and exceeds GPT-3 by over 8\% point, indicating that ERNIE 3.0 Titan is superior in learning and reasoning.

\subsubsection{Case Study}\label{sec: zero-shot-case-study}

\begin{table}[]
\centering
\resizebox{\textwidth}{!}{%
\begin{tabular}{@{}llccccc@{}}
\toprule \toprule
\textbf{Type}                       & \textbf{Task (\# of cases)}              & \textbf{CPM-1}       & \textbf{PLUG}        & \textbf{PanGu-$\alpha$} & \textbf{ERNIE 3.0}   & \textbf{ERNIE 3.0 Titan} \\ \midrule
\multirow{3}{*}{Question Answering} & Factual QA (30)                          & 1.67/1.50/1.03       & 1.23/0.83/0.27       & 1.60/1.07/0.60          & 1.67/1.50/1.03       & \textbf{1.97/1.77/1.30}  \\
                                    & Opinion QA (30)                          & 1.27/0.80/-          & 1.43/1.13/-          & 1.60/1.23/-             & 1.67/1.33/-          & \textbf{1.70/1.73/-}     \\
                                    & Reasoning  (30)                          & 1.20/0.83/0.27       & 1.03/0.83/0.07       & 1.03/0.83/0.00          & 1.70/1.60/0.23       & \textbf{1.90/1.93/0.37}  \\ \midrule
\multirow{2}{*}{Interpretation}     & Interpretation of Terms (30)             & 1.23/0.73/0.70       & 1.50/0.97/0.80       & 1.57/0.97/0.70          & 1.83/1.60/1.33       & \textbf{1.87/1.87/1.53}  \\
                                    & Reverse Dictionary (28)                  & 0.11/0.11/0.07       & 1/0.86/0.36          & 1.32/1.00/1.00          & 1.43/1.32/0.93       & \textbf{1.64/1.36/1.14}  \\ \midrule
\multirow{2}{*}{Dialogue}           & Single-Turn Dialogue (50)                & 1.50/0.86/-          & 1.38/0.12/-          & 1.46/0.78/-             & \textbf{1.84}/0.58/- & 1.74/\textbf{1.46}/-     \\
                                    & Multi-Turn Dialogue (50)                 & 1.10/0.83/-          & 0.90/0.62/-          & 1.02/0.96/-             & \textbf{1.50}/1.14/- & 1.44/\textbf{1.54}/-     \\ \midrule
\multirow{5}{*}{Text Generation}    & Recipe Generation (30)                   & 0.80/0.63/-          & \textbf{1.67}/1.03/- & 1.40/1.03/-             & 1.30/1.10/-          & 1.50/\textbf{1.13}/-     \\
                                    & Novel Generation (50)                    & 0.92/0.86/-          & 1.24/0.94/-          & 1.34/\textbf{1.06}/-    & 1.27/1.04/-          & \textbf{1.44/1.06/-}     \\
                                    & Professional Manuscripts Generation (50) & 1.04/0.82/-          & 1.48/1.06/-          & 1.36/0.86/-             & 1.32/1.10/-          & \textbf{1.74/1.48}       \\
                                    & Couplet Generation (30)                  & 0.73/0.60/-          & 0.77/0.86/-          & 1.10/0.90/-             & 1.50/1.47/-          & \textbf{1.83/1.73/-}     \\
                                    & Poetry Generation (30)                   & 1.80/\textbf{1.60}/- & 1.17/1.00/-          & 1.833/1.07/-            & \textbf{1.87}/1.30/- & 1.73/1.56/-              \\ \midrule
Summarization                       & News Summarization  (29)         & 1.21/1.10/-          & 0.93/0.86/-          & 1.24/1.03/-             & 1.41/1.31/-          & \textbf{1.52/1.34/-}     \\ \midrule
\multicolumn{2}{l}{\textbf{Average}}                                           & 1.09/0.81/0.38       & 1.21/0.86/0.37       & 1.38/0.98/0.58          & 1.56/1.26/0.88       & \textbf{1.69/1.53/1.09}  \\ \bottomrule \bottomrule
\end{tabular}%
}
\\[5pt]
\caption{The zero-shot generation performance manually evaluated on our collected 467 cases. (we reported the average score of \textit{coherence}, \textit{fluency}, and \textit{accuracy} respectively on a scale of
{[}0, 1, 2{]})}
\label{tab:zero-shot-generation}
\end{table}

\begin{table}[]
\centering
\resizebox{0.7\textwidth}{!}{%
\begin{tabular}{@{}llcc@{}}
\toprule \toprule
\textbf{Type}                       & \textbf{Task (\# of cases)}              & \textbf{GPT-3}       & \textbf{ERNIE 3.0 Titan} \\ \midrule
\multirow{3}{*}{Question Answering} & Factual QA (11)                          & 1.73/1.64/1.36       & \textbf{2.00/1.73/1.55}  \\
                                    & Opinion QA (29)                          & 1.07/1.03/-          & \textbf{1.69/1.72/-}     \\
                                    & Reasoning  (30)                          & 1.40/1.30/0.30       & \textbf{1.90/1.93/0.37}  \\ \midrule
\multirow{2}{*}{Interpretation}     & Interpretation of Terms (14)             & 1.00/0.86/0.57       & \textbf{1.71/1.93/1.29}  \\
                                    & Reverse Dictionary (28)                  & 0.71/0.68/0.64       & \textbf{1.64/1.36/1.14}  \\ \midrule
\multirow{2}{*}{Dialogue}           & Single-Turn Dialogue (50)                & 0.82/0.46/-          & 1.74/\textbf{1.46}/-     \\
                                    & Multi-Turn Dialogue (50)                 & 0.94/0.64/-          & \textbf{1.44/1.54/-}     \\ \midrule
Text Generation                     & Recipe Generation (5)                    & 1.00/\textbf{1.00}/- & \textbf{1.40/1.00/-}     \\
                                    & Novel Generation (50)                    & 1.08/0.80/-          & \textbf{1.44/1.06/-}     \\
                                    & Professional Manuscripts Generation (50) & 1.54/1.34/-          & \textbf{1.76/1.48}       \\ \midrule
Summarization                       & News Summarization  (29)                 & 0.46/0.36/-          & \textbf{1.57/1.32/-}     \\ \midrule
\multicolumn{2}{l}{\textbf{Average}}                                           & 1.07/0.92/0.72       & \textbf{1.66/1.50/1.09}  \\ \bottomrule \bottomrule
\end{tabular}%
}
\caption{The zero-shot generation performance between GPT-3 and our model on a subset of manually collected cases. (Cases requiring background knowledge about Chinese history, geography, and culture have been removed. We reported the average score of \textit{coherence}, \textit{fluency}, and \textit{accuracy} respectively on a scale of {[}0, 1, 2{]}).}
\label{tab:zero-shot-generation-gpt3}
\end{table}

\begin{table}[]
\centering
\resizebox{\textwidth}{!}{%
\begin{tabular}{@{}ll@{}}
\toprule \toprule
\textbf{Input (Genre:News)}                 & \begin{CJK}{UTF8}{gbsn}[News-0] [News-1] ... [News-63]\end{CJK}                                                                                                                                                    \\ \cmidrule(l){2-2}
\textbf{Generation}                 & \parbox{0.8\textwidth}{\begin{CJK}{UTF8}{gbsn} 一、二级公立医疗服务体系建设的关键是建立起有效率的服务体系,即能够满足群众的需要,而不是单纯的追求最大的经济利益。因此要重视医德
和业务素质的培养,要求医生要有爱心,要有责任心,要有耐心。\\
The key to the construction of the first and second level public medical service system is to establish an efficient service system, that is, to meet the needs of the public, rather than simply pursuing the greatest economic benefits. Therefore, we should pay attention to the training of medical ethics and business quality, requiring doctors to be caring, responsible, and patient.
\end{CJK}}              \\ \midrule
\textbf{Input (Genre:QA)}         & \begin{CJK}{UTF8}{gbsn}[QA-0] [QA-1] ... [QA-63]\end{CJK}                                                                                                                                                      \\ \cmidrule(l){2-2} 
\textbf{Generation}                 & \parbox{0.8\textwidth}{\begin{CJK}{UTF8}{gbsn}如何让手机的图片变得好看？1.美图秀秀软件 2.photoshop软件 3.snapseed软件\\
How to make the pictures on your phone look good? 1. Meitu software 2. Photoshop software 3. Snapseed software
\end{CJK}}                         \\ \midrule
\textbf{Input (Genre:Novel)} & \begin{CJK}{UTF8}{gbsn}[Novel-0][Novel-1] ... [Novel-63]\end{CJK}                                                                                                                                                 \\ \cmidrule(l){2-2} 
\textbf{Generation}                 & \parbox{0.8\textwidth}{\begin{CJK}{UTF8}{gbsn}“不过是一个人，还不至于如此小心。”李青阳看着手中的那颗丹药说道。\\ 
"It's just one person, no need to be so careful." Qingyang Li looked at the pill in his hand and said.
\end{CJK}}                                                                                                                                                      \\ \midrule
\textbf{Input (Topic:Digital)}   & \parbox{0.8\textwidth}{\begin{CJK}{UTF8}{gbsn}[t] 数码 [/t]\end{CJK}}                                           \\ \cmidrule(l){2-2} 
\textbf{Generation}                 & \parbox{0.8\textwidth}{\begin{CJK}{UTF8}{gbsn}iphone12pro max可以插两张卡吗? 可以。\\ Does iPhone 12 Pro Max support Dual SIM? Yes. \end{CJK}}                                  \\ \midrule
\textbf{Input (Topic:Astrology Horoscope)} & \begin{CJK}{UTF8}{gbsn}[t] 星座运势 [/t]\end{CJK}                                                                                                                                                  \\ \cmidrule(l){2-2} 
\textbf{Generation}                 & \parbox{0.8\textwidth}{\begin{CJK}{UTF8}{gbsn}天秤座的人，在12月底，财运上上签，运势如虹，财源广进。\\ The Libra, at the end of December, is on a financial upswing, with a wealthy and prosperous fortune. \end{CJK}}                                                                                                                                                         \\ \midrule
\textbf{Input (Keywords:Desert)} & \begin{CJK}{UTF8}{gbsn}[k] 沙漠 [/k]\end{CJK}                                                                                                                                                  \\ \cmidrule(l){2-2} 
\textbf{Generation}                 & \parbox{0.8\textwidth}{\begin{CJK}{UTF8}{gbsn}沙漠之心在我的世界里面可是一个稀有物品啊，那可是只有在沙漠中才能找到。\\ The Desert Heart is a rare item in the Minecraft, which can only be found in the desert. \end{CJK}}                                                                                                                                                         \\ \midrule
\textbf{Input (Keywords:Beijing, Winter)} & \begin{CJK}{UTF8}{gbsn}[k] 北京,冬天 [/k]\end{CJK}                                                                                                                                                  \\ \cmidrule(l){2-2} 
\textbf{Generation}                 & \parbox{0.8\textwidth}{\begin{CJK}{UTF8}{gbsn}北京的冬天太长了，这个星期都没有太阳，还不能晒被子，只能在宿舍里面躺着。\\ The winter in Beijing is too long, it's cloudy this week. You can't hang out the quilts, so just lounge around in the dorm.\end{CJK}}                                                                                                                                                         \\ \midrule
\textbf{Input (Genre:QA; Keywords:Beijing, Winter)} & \begin{CJK}{UTF8}{gbsn}[QA-0] [QA-1] ... [QA-63] [k] 北京,冬天 [/k]\end{CJK}                                                                                                                                                  \\ \cmidrule(l){2-2} 
\textbf{Generation}                 & \parbox{0.8\textwidth}{\begin{CJK}{UTF8}{gbsn}北京冬天有哪些好吃的美食推荐呢？北京的冬天最好吃的当然是涮羊肉啦，羊肉的营养价值可是很高的。\\ What are some of the best winter food recommendations in Beijing? The best thing to eat in Beijing in winter is instant-boiled mutton slices, which has a very high nutritional value.\end{CJK}}                                                                                                                                                         \\ \midrule
\textbf{Input (Genre:Novel; Keywords:Beijing, Winter)} & \begin{CJK}{UTF8}{gbsn} [Novel-0][Novel-1] ... [Novel-63] [k] 北京,冬天 [/k] \end{CJK}                                                                                                                                                  \\ \cmidrule(l){2-2} 
\textbf{Generation}                 & \parbox{0.8\textwidth}{\begin{CJK}{UTF8}{gbsn}北平的冬天很干燥，尤其是在城南，城南的一户人家，门口放着一架风车，风车吱吱地转着，风里透着一股冷气。\\ It' s a dry winter in Peking, especially in the south of the city. In a house in the south of the city, a windmill was placed at the door, and the windmill creaked and turned with a cold wind.\end{CJK}}                                                                                                                                                         \\ \midrule
\textbf{Input (Genre:Novel; Topic:Pet; Keywords:Beijing, Winter)} & \parbox{0.8\textwidth}{\begin{CJK}{UTF8}{gbsn} [QA-0] [QA-1] ... [QA-63] [t] 宠物 [/t] [k] 北京,冬天 [/k] \end{CJK}} \\ \cmidrule(l){2-2} 
\textbf{Generation}                 & \parbox{0.8\textwidth}{\parbox{0.8\textwidth}{\begin{CJK}{UTF8}{gbsn}北京哪里可以找一只狗做宠物，冬天太难熬，想找一只狗陪我度过寒冬。\\ Where to find a dog as a pet in Beijing, it' s too hard to survive the winter, I want to find a dog to spend the winter with me.\end{CJK}}}                                                                     \\ \bottomrule \bottomrule
\end{tabular}%
}
\\[5pt]
\caption{Illustrations of controllable generations from ERNIE 3.0 Titan.}
\label{tab:gen-examples}
\end{table}

We manually collected 467 cases~\footnote{\url{https://ernie-github.cdn.bcebos.com/cases.xlsx}} to evaluate the zero-shot generation ability of current large-scale pre-trained models on 13 tasks from 5 different types including Question Answering, Interpretation, Dialogue, Text Generation and Summarization. In human evaluation, the annotators are asked to score the generation quality on a scale of $[0, 1, 2]$. We reported the average score of \textit{coherence}, \textit{fluency}, and \textit{accuracy} in Tab.~\ref{tab:zero-shot-generation},~\ref{tab:zero-shot-generation-gpt3}, and showed some controllable generations of ERNIE 3.0 Titan in Tab.~\ref{tab:gen-examples}. In addition, we construct a subset of above manually collected cases where cases requiring background knowledge about Chinese history, geography, and culture have been removed. Overall, ERNIE 3.0 Titan can generate the most coherent, fluent and accurate texts on average as compared to CPM-1, PLUG, PanGu-$\alpha$, ERNIE 3.0 and GPT-3~\footnote{We use the implementation of CPM-1 in \url{https://github.com/jm12138/CPM-Generate-Paddle}, PLUG in \url{https://nlp.aliyun.com/portal?/BigText_chinese\#/BigText_chinese}, PanGu-$\alpha$ in \url{https://git.openi.org.cn/PCL-Platform.Intelligence/PanGu-Alpha}, ERNIE 3.0 in \url{https://wenxin.baidu.com/wenxin/ernie} and GPT-3 in \url{https://beta.openai.com/docs/guides/completion}}, and users can combine different attributes to generate highly credible and controllable generations. The introduction of three scoring metrics are listed as follows, and the scoring details are provided in Tab.~\ref{tab:score-detail}.
\begin{itemize}
    \item \textbf{Coherence} measures whether the generation is relevant and consistent with the context.
    \item \textbf{Fluency} evaluates whether the generated text is natural or readable. A fluent text should have no semantic contradiction among the generated text.
    \item \textbf{Accuracy} is a metric to evaluate whether the generated text is the same as the ground truth.
\end{itemize}

\begin{table}[]
\centering
\resizebox{\textwidth}{!}{%
\begin{tabular}{@{}c|lll@{}}
\toprule \toprule
\textbf{Score} & \textbf{Coherence}                                                                                                                                    & \textbf{Fluency}                                                                                                                                 & \textbf{Accuracy}             \\ \midrule
0              & \begin{tabular}[c]{@{}l@{}}The generation is not related to the context.\\ The generation has obvious conflicts with the context.\end{tabular}        & \begin{tabular}[c]{@{}l@{}}The generation is unnatural.\\ There are contradictions in the generated text.\end{tabular}                           & The answer is wrong.          \\ \midrule
1              & \begin{tabular}[c]{@{}l@{}}The generation is weakly related to the context.\\ The generation has minor logic conflicts with the context.\end{tabular} & \begin{tabular}[c]{@{}l@{}}The generation has minor influent part.\\ The generation slightly influences the reading.\end{tabular}                & The answer is partly correct. \\ \midrule
2              & \begin{tabular}[c]{@{}l@{}}The generation is strongly related to the context.\\ The logic in the generation is aligned with the context.\end{tabular} & \begin{tabular}[c]{@{}l@{}}The generation is semantically complete and fluent.\\ There are no contradictions in the generated text.\end{tabular} & The answer is correct.        \\ \bottomrule \bottomrule
\end{tabular}%
}
\\[5pt]
\caption{Scoring details for zero-shot generation.}
\label{tab:score-detail}
\end{table}

\subsection{Experiments on Model Distillation}

This section discusses the experiment for our task-agnostic model distillation framework for ERNIE 3.0 Titan. We evaluate our distilled ERNIE 3.0 Titan on five typical types of downstream tasks, including natural language inference (XNLI~\cite{conneau2018xnli}), semantic analysis (ChnSentiCorp~\cite{song6chnsenticorp}), document question answering (NLPCC-DBQA\footnote{http://tcci.ccf.org.cn/conference/2016/dldoc/evagline2.pdf}), semantic similarity (LCQMC~\cite{liu2018lcqmc}), and machine reading comprehension (CMRC2018~\cite{cmrc2018}).
 
% introduce exp setting
To reduce the gap between the giant teacher model and the student models, we introduce a 24-layers TA model with a hidden size of 1024. To accelerate the distillation procedure, we also pre-train the TA model for 500K steps before starting distillation with the same pre-training setting as ERNIE 3.0 Titan. 
For ALD settings, we add one extra layer for each student. Then we jointly train the teacher and students with OFD.
We introduce five students with different model sizes and parameters ranging from 14M to 110M. 
We measure the inference latency of these student models on a V100 GPU with PaddlePaddle and show the relative speedup with respect to BERT-Base in Table~\ref{tab:dist-exp}.

\begin{wrapfigure}{r}{0.4\textwidth}
    \begin{center}
    \includegraphics[width=0.38\textwidth]{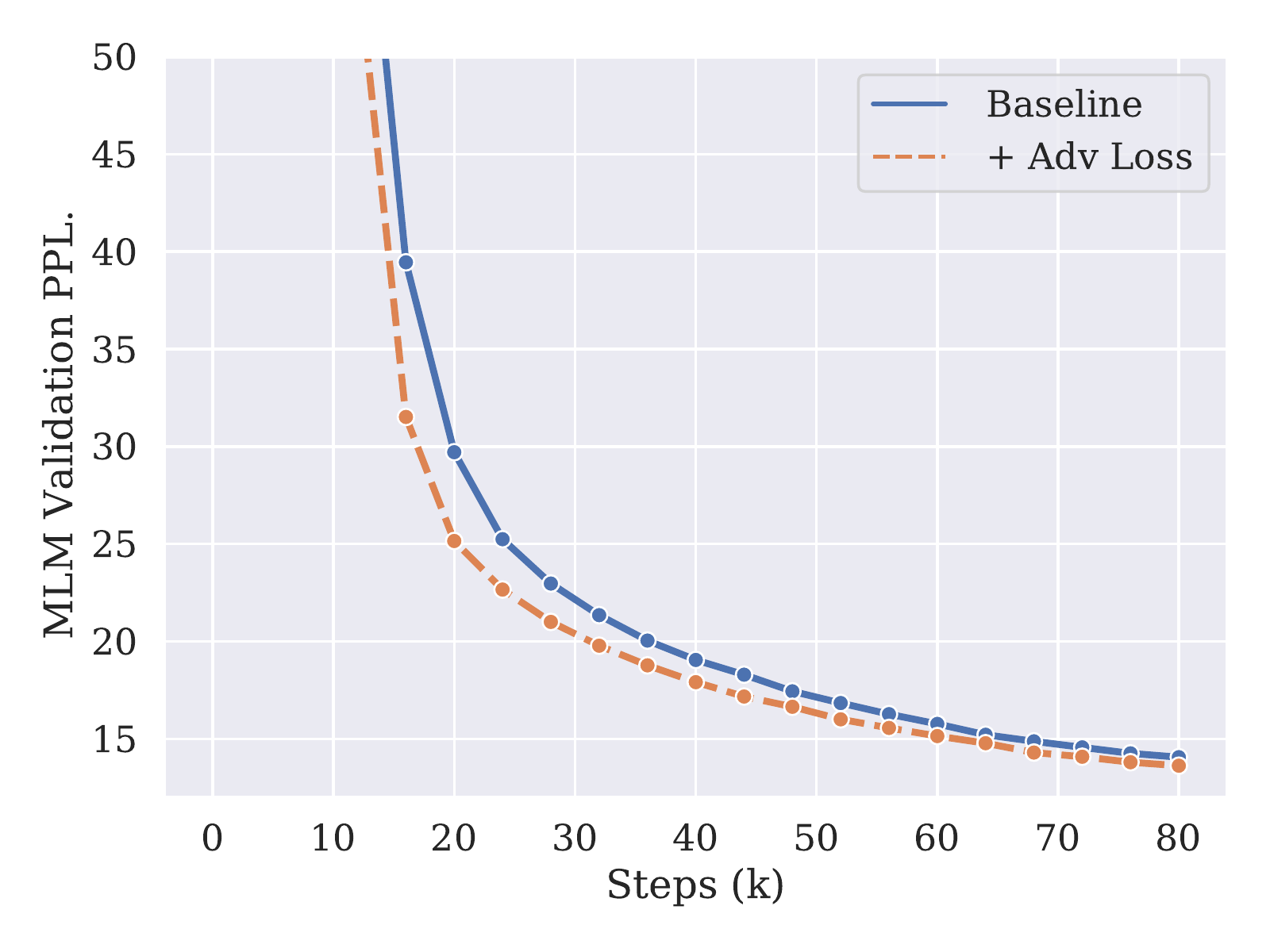}
    \end{center}
    \caption{Perplexity variation of the mask language model task with respect to training steps.}
    \label{fig:ablation-adv-loss}
\end{wrapfigure}

We compare our downstream fine-tuning results with other compact PLMs.
% introduce baseline
We choose the Chinese version BERT~\cite{devlin2018bert}, TinyBERT~\cite{tinybert}, ERNIE 2.0~\cite{sun2020ernie}, and RoBERTa-wwm-ext~\cite{liu2019roberta} as baseline models. BERT, RoBERTa-wwm-ext and ERNIE 2.0 are PLMs pre-trained from scratch without any distillation. TinyBERT is pre-trained via a multi-step distillation procedure using the task-specific distillation paradigm. Results are shown in Table~\ref{tab:dist-exp}, the experiments are reported with means over five random initialization. 

% show results
As shown in Table~\ref{tab:dist-exp}, the 12L768H student model achieves SOTA results on all tasks. By taking a closer look at the EM and F1 for CMRC2018, this student model surpasses the strong baseline ERNIE 2.0 by 4.78 and 2.82, respectively. Notably, among all 6-layer PLMs, the 6L768H version of distilled ERNIE 3.0 Titan performs the best and even outperforms the 12-layer BERT-Base on XNLI, LCQMC, and NLPCC-DBQA. Comparison between these 6-layer students and TinyBERT also demonstrate the effectiveness of our proposed framework.

\begin{table}[]
\resizebox{\textwidth}{!}{%

\begin{tabular}{cccccccccccccc}
\toprule \toprule
\multirow{3}{*}{Model}                                                                         & \multirow{3}{*}{Arch.} & \multirow{3}{*}{Speedup} & \multicolumn{2}{c}{Natural Language Inference} & \multicolumn{2}{c}{Sentiment Analysis} & \multicolumn{2}{c}{Document QA}             & \multicolumn{2}{c}{Semantic Similarity} & \multicolumn{2}{c}{Reading Comprehension} & \multirow{3}{*}{Avg.} \\
                                                                                               &                       &                           & \multicolumn{2}{c}{XNLI}                       & \multicolumn{2}{c}{ChnSentiCorp}                & \multicolumn{2}{c}{NLPCC-DBQA}                    & \multicolumn{2}{c}{LCQMC}               & \multicolumn{2}{c}{CMRC2018}                  &                       \\
                                                                                               &                       &                           & Dev                    & Test                  & Dev                & Test              & Dev            & Test                       & Dev                & Test               & EM                  & F1                  &                       \\ \hline
BERT-Base                                                                                      & 12L768H               & 1.0x                      & 77.28                  & 77.14                 & 94.94              & 95.58             & 78.66          & 80.62          & 86.95              & 85.53              & 67.00               & 86.31               & 83.11                 \\
RoBERTa-wwm-ext-Base                                                                           & 12L768H               & 1.0x                      & 79.20                  & 78.30                 & 94.60              & 94.80             & 83.34          & 83.71          & 88.70              & 86.10              & 66.50               & 86.50               & 83.88                 \\
ERNIE 2.0 Base                                                                                 & 12L768H               & 1.0x                      & 81.20            & 79.70           & \textbf{95.70}     & 95.50             & 84.70    & 85.30    & 90.90        & 87.90        & 69.10               & 88.60               & 85.45                 \\ \hline
TinyBERT                                                                                       & 6L768H                & 2.0x                      & 76.36                  & 75.48                 & 93.64              & 95.03             & 78.80          & 80.00          & 87.40              & 85.95              & 65.01               & 84.78               & 82.27                 \\
RoBERTa-wwm-ext-6L                                                                             & 6L768H                & 2.0x                      & 77.88                  & 76.44                 & 94.78              & 94.38             & 80.59          & 81.60          & 88.31              & 86.30              & 65.75               & 85.71               & 82.89                 \\ \hline
\multirow{5}{*}{\begin{tabular}[c]{@{}c@{}}ERNIE 3.0 Titan\\ Online Distillation\end{tabular}} & 12L768H               & 1.0x                      & \textbf{83.37}         & \textbf{81.88}        & {95.46}        & \textbf{96.13}    & \textbf{85.80} & \textbf{86.88} & \textbf{90.98}     & \textbf{88.84}     & \textbf{73.88}      & \textbf{91.42}      & \textbf{87.28}        \\
                                                                                               & 6L768H                & 2.0x                      & 80.28                  & 79.55                 & 95.14              & 95.77       & 83.60          & 84.54          & 88.99              & 87.53              & 71.98         & 89.37         & 85.61           \\
                                                                                               & 6L384H                & 4.8x                      & 76.38                  & 75.94                 & 94.12              & 94.45             & 79.72          & 81.44          & 87.66              & 86.13              & 61.58               & 83.29               & 82.08                 \\
                                                                                               & 4L384H                & 7.0x                      & 74.48                  & 73.47                 & 93.32              & 93.45             & 77.73          & 79.56          & 86.79              & 85.77              & 59.13               & 81.58               & 80.52                 \\
                                                                                               & 4L312H                & 8.5x                      & 74.03                  & 73.60                 & 93.53              & 94.03             & 77.22          & 78.22          & 86.13              & 86.38              & 58.71               & 80.78               & 80.40                 \\ 
 \bottomrule \bottomrule
\end{tabular}
}
\\[5pt]
\caption{Results for model compression on Chinese tasks.}
\label{tab:dist-exp}
\end{table}

\subsection{Analysis}

\subsubsection{The Effect of Different Scoring Functions}\label{sec: analysis-scoring-function}

In Table.~\ref{tab:ablation-zero-shot}, we compare the performance of different scoring functions on five zero-shot tasks. Overall, $P(y|x)/P(y)$-\textit{bi} is more amenable to text classification tasks, while semantic similarity and natural language inference tasks prefer $P(\text{True}|x, y)$-\textit{bi}. We put forward a hypothesis that the effectiveness of $P(y|x)/P(y)$-\textit{bi} is positively correlated with the size of the label set. On IFLYEK-ZC with 119 classes, ERNIE 3.0 Titan obtains 9.68\% point ($30.74 \rightarrow 40.42$) performance gain compared to 4.28\% point ($53.55 \rightarrow 57.83$) on TNEWS-FC with 15 classed when eliminating the label bias effect through the division of $P(y)$. $P(y|x)/P(y)$-\textit{bi} even fails on BUSTM with 2 classes and OCNLI-FC with 3 classes compared to the second-best scoring function $P(x|y)$-\textit{uni}. Intuitively, the dataset with a larger label set is more likely to be affected by the label bias. The model prefers frequent answers in the pre-training dataset, conflicting with the balanced label distribution in the downstream dataset. $P(\text{True}|x, y)$-\textit{bi} utilizes the inherently pre-trained NSP-task, which is suitable for tasks that need to distinguish the semantic similarity between two sentences. The best performance always achieves using the scoring function with bidirectional attention. When comparing $P(y|x)$-\textit{uni} with $P(y|x)$-\textit{bi}, $P(y|x)$-\textit{bi} could additionally utilize the information from the text behind the label resulting in better performance among all datasets.

\begin{table}[]
\centering
\resizebox{0.7\textwidth}{!}{%
\begin{tabular}{@{}lccccc@{}}
\toprule \toprule
\textbf{Scoring Function\&Dataset} & \textbf{IFLYEK-ZC} & \textbf{TNEWS-FC} & \textbf{BUSTM}    & \textbf{OCNLI-FC} & \textbf{CHID-FC}     \\ \midrule
P(x, y)-\textit{uni}               & 19.16              & 40.71             & 51.88             & 38.13             & \underline{87.12} \\
P(x|y)-\textit{uni}                & 27.90              & 52.91             & \underline{57.50} & \underline{46.88} & 54.95             \\
P(y|x)-\textit{uni}                & 25.78            & 38.07             & 48.75             & 32.50             & 79.70             \\
P(y|x)/P(y)-\textit{uni}           & \underline{36.13}            & \underline{56.01} & 48.75             & 38.13             & 76.24             \\ \midrule
P(x|y)-\textit{bi}                 & 31.83              & 47.09             & 48.13             & 37.50             & 45.05             \\
P(y|x)-\textit{bi}                 & 30.74              & 53.55             & 51.25             & 33.75             & \textbf{88.61}    \\
P(y|x)/P(y)-\textit{bi}            & \textbf{40.42}     & \textbf{57.83}    & 51.25             & 35.63             & 84.16             \\ \midrule
P(True|x, y)-\textit{bi}           & 30.52              & 49.18             & \textbf{64.37}    & \textbf{53.75}    & 23.27             \\ \bottomrule \bottomrule
\end{tabular}%
}
\\[5pt]
\caption{Comparison of different scoring functions using ERNIE 3.0 Titan on zero-shot learning tasks. \textit{uni}, \textit{bi} means the unidirectional and bi-directional attention used to calculate the score.}
\label{tab:ablation-zero-shot}
\end{table}

\subsubsection{Adversarial Credibility Classification}\label{sec: adv}

We notice that the adversarial credibility classification method mentioned in \ref{sec: pretrain-task} not only can filter out low credibility texts during the generation but also can speed up the pre-training convergence. We conduct a comparative experiment with the base model settings (12 layers, 768 dims, 12 attention heads) and report the results in Figure \ref{fig:ablation-adv-loss}. The pre-training tasks of our baseline strictly follow the settings of the original ERNIE 3.0, while the contrast model has an additional self-supervised adversarial loss for credibility classification.

Figure \ref{fig:ablation-adv-loss} illustrates the perplexity variation of the masked language model task during the pre-training process. The model with the auxiliary adversarial loss reaches a higher convergence speed. We think this might because the adversarial loss of distinguishing whether a text is generated or the original one can help the model learn the distribution of true natural sentences.

\subsubsection{The Effect of Controllable Attributes}

\begin{wrapfigure}{r}{0.4\textwidth}
    \begin{center}
    \includegraphics[width=0.4\textwidth]{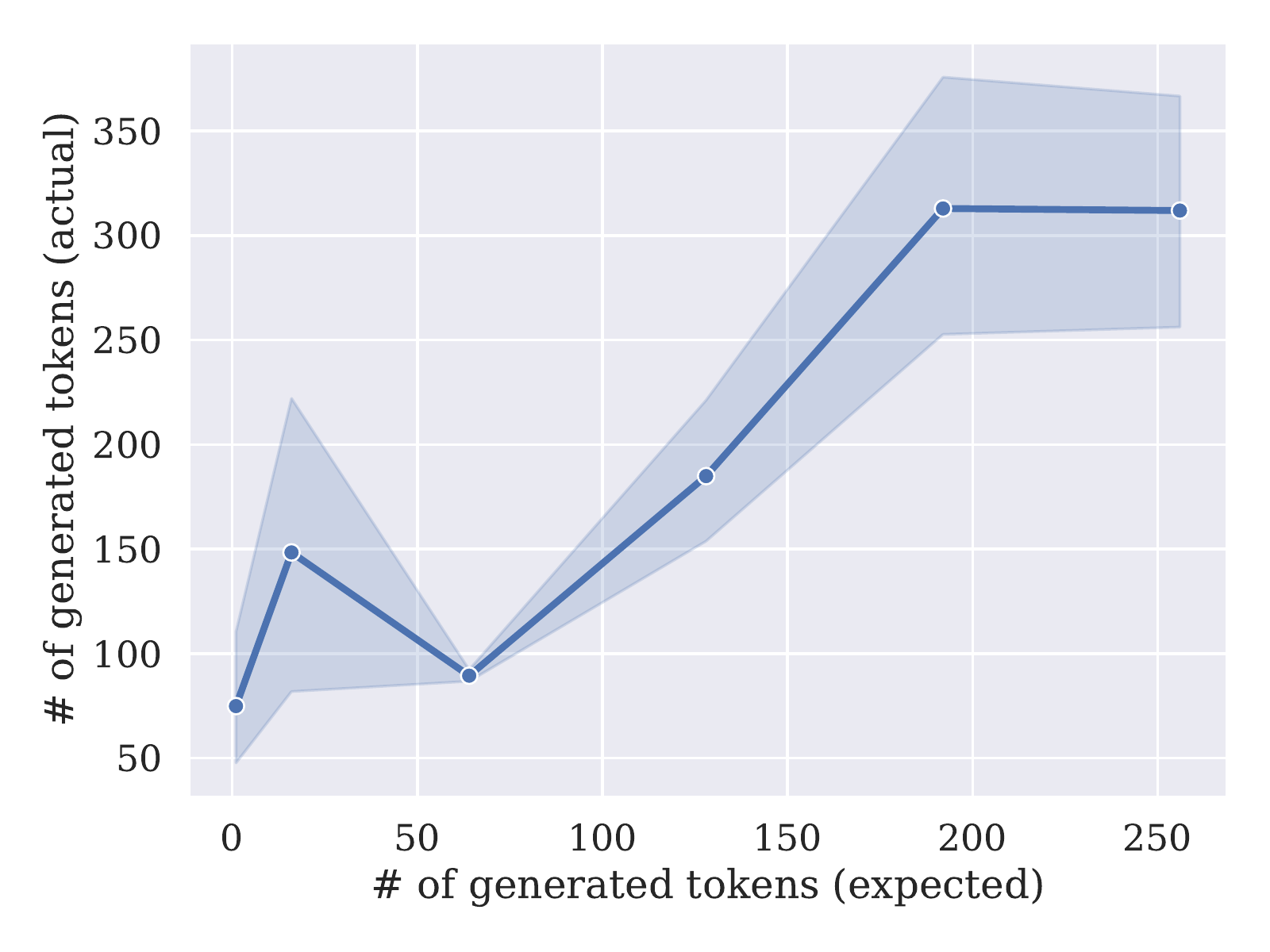}
    \end{center}
    \caption{The number of generated tokens controlled by the length attribute.}
    \label{fig:word-length}
\end{wrapfigure}

In ERNIE 3.0 Titan framework, we have introduced five different controllable attributes, including genre, topic, keywords, sentiment, and length in Sec.~\ref{sec:pre-training data}. By assembling different attributes, users can have more access to ERNIE 3.0 Titan to obtain diverse and controllable generations shown in Table.~\ref{tab:gen-examples}. Meanwhile, the genre soft prompts improve the performance on some zero-shot tasks such as CHID, CMNLI, and WPLC. We assume the improvement comes from the domain calibration. For example, \texttt{[novelN]} soft prompts are prefix of novel texts when optimizing the language modeling loss. When conditioning on the \texttt{[novelN]} soft prompts, the output distribution of the model will be shifted to the novel domain which results in the lower perplexity score on WPLC dataset. In addition, we test the effect of the length attribute showing that the number of actual generated tokens has a positive correlation with the expected generated tokens in Figure.~\ref{fig:word-length}. Also, we observe that the length attribute affects the genre of generations. ERNIE 3.0 Titan tends to generate texts constructed from knowledge graph (like \texttt{The capital of China is Beijing.}) when the length attribute is small and prefers novel and web texts when the length attribute is large. 

\section{Conclusion} 
We pre-train a knowledge-enhanced language model with 260 billion parameters named ERNIE 3.0 Titan based on the ERNIE 3.0 framework. It is the largest Chinese dense pre-training model as far as we know. We have validated it on 68 datasets, and the results show that ERNIE 3.0 Titan achieves new state-of-the-art results. In addition, We propose a novel method for users to control the generation result and obtain the result factually consistent with the real world. We also devise an online distillation framework and conduct several distilled models of different sizes concerning the computation overhead of large-scale pre-training models. In the next stage, we will continually update ERNIE 3.0 Titan with more data to further explore the limit of the performance of large-scale pre-trained language models. We will also endeavor to explore the potential of knowledge-enhanced large-scale multi-modal models for more and various tasks.

\bibliographystyle{unsrt}  
\bibliography{references}  

\begin{thebibliography}{100}

\bibitem{gpt-3}
Tom Brown, Benjamin Mann, Nick Ryder, Melanie Subbiah, Jared~D Kaplan, Prafulla
  Dhariwal, Arvind Neelakantan, Pranav Shyam, Girish Sastry, Amanda Askell,
  Sandhini Agarwal, Ariel Herbert-Voss, Gretchen Krueger, Tom Henighan, Rewon
  Child, Aditya Ramesh, Daniel Ziegler, Jeffrey Wu, Clemens Winter, Chris
  Hesse, Mark Chen, Eric Sigler, Mateusz Litwin, Scott Gray, Benjamin Chess,
  Jack Clark, Christopher Berner, Sam McCandlish, Alec Radford, Ilya Sutskever,
  and Dario Amodei.
\newblock Language models are few-shot learners.
\newblock In H.~Larochelle, M.~Ranzato, R.~Hadsell, M.~F. Balcan, and H.~Lin,
  editors, {\em Advances in Neural Information Processing Systems}, volume~33,
  pages 1877--1901. Curran Associates, Inc., 2020.

\bibitem{sun2021ernie}
Yu~Sun, Shuohuan Wang, Shikun Feng, Siyu Ding, Chao Pang, Junyuan Shang,
  Jiaxiang Liu, Xuyi Chen, Yanbin Zhao, Yuxiang Lu, Weixin Liu, Zhihua Wu,
  Weibao Gong, Jianzhong Liang, Zhizhou Shang, Peng Sun, Wei Liu, Xuan Ouyang,
  Dianhai Yu, Hao Tian, Hua Wu, and Haifeng Wang.
\newblock Ernie 3.0: Large-scale knowledge enhanced pre-training for language
  understanding and generation.
\newblock {\em arXiv preprint arXiv:2107.02137}, 2021.

\bibitem{ma2019paddlepaddle}
Yanjun Ma, Dianhai Yu, Tian Wu, and Haifeng Wang.
\newblock Paddlepaddle: An open-source deep learning platform from industrial
  practice.
\newblock {\em Frontiers of Data and Domputing}, 1(1):105--115, 2019.

\bibitem{peters2018deep}
Matthew~E Peters, Mark Neumann, Mohit Iyyer, Matt Gardner, Christopher Clark,
  Kenton Lee, and Luke Zettlemoyer.
\newblock Deep contextualized word representations.
\newblock {\em arXiv preprint arXiv:1802.05365}, 2018.

\bibitem{radford2018improving}
Alec Radford, Karthik Narasimhan, Tim Salimans, and Ilya Sutskever.
\newblock Improving language understanding by generative pre-training.
\newblock {\em URL https://s3-us-west-2. amazonaws.
  com/openai-assets/research-covers/languageunsupervised/language understanding
  paper. pdf}, 2018.

\bibitem{devlin2018bert}
Jacob Devlin, Ming-Wei Chang, Kenton Lee, and Kristina Toutanova.
\newblock Bert: Pre-training of deep bidirectional transformers for language
  understanding.
\newblock {\em arXiv preprint arXiv:1810.04805}, 2018.

\bibitem{sun2019ernie}
Yu~Sun, Shuohuan Wang, Yukun Li, Shikun Feng, Xuyi Chen, Han Zhang, Xin Tian,
  Danxiang Zhu, Hao Tian, and Hua Wu.
\newblock Ernie: Enhanced representation through knowledge integration.
\newblock {\em arXiv preprint arXiv:1904.09223}, 2019.

\bibitem{tinybert}
Xiaoqi Jiao, Yichun Yin, Lifeng Shang, Xin Jiang, Xiao Chen, Linlin Li, Fang
  Wang, and Qun Liu.
\newblock Tinybert: Distilling {BERT} for natural language understanding.
\newblock In Trevor Cohn, Yulan He, and Yang Liu, editors, {\em Findings of the
  Association for Computational Linguistics: {EMNLP} 2020, Online Event, 16-20
  November 2020}, volume {EMNLP} 2020 of {\em Findings of {ACL}}, pages
  4163--4174. Association for Computational Linguistics, 2020.

\bibitem{minilm}
Wenhui Wang, Furu Wei, Li~Dong, Hangbo Bao, Nan Yang, and Ming Zhou.
\newblock Minilm: Deep self-attention distillation for task-agnostic
  compression of pre-trained transformers.
\newblock In Hugo Larochelle, Marc'Aurelio Ranzato, Raia Hadsell,
  Maria{-}Florina Balcan, and Hsuan{-}Tien Lin, editors, {\em Advances in
  Neural Information Processing Systems 33: Annual Conference on Neural
  Information Processing Systems 2020, NeurIPS 2020, December 6-12, 2020,
  virtual}, 2020.

\bibitem{minilmv2}
Wenhui Wang, Hangbo Bao, Shaohan Huang, Li~Dong, and Furu Wei.
\newblock Minilmv2: Multi-head self-attention relation distillation for
  compressing pretrained transformers.
\newblock In Chengqing Zong, Fei Xia, Wenjie Li, and Roberto Navigli, editors,
  {\em Findings of the Association for Computational Linguistics: {ACL/IJCNLP}
  2021, Online Event, August 1-6, 2021}, volume {ACL/IJCNLP} 2021 of {\em
  Findings of {ACL}}, pages 2140--2151. Association for Computational
  Linguistics, 2021.

\bibitem{ernie-tiny}
Weiyue Su, Xuyi Chen, Shikun Feng, Jiaxiang Liu, Weixin Liu, Yu~Sun, Hao Tian,
  Hua Wu, and Haifeng Wang.
\newblock Ernie-tiny : {A} progressive distillation framework for pretrained
  transformer compression.
\newblock {\em CoRR}, abs/2106.02241, 2021.

\bibitem{carbon}
David~A. Patterson, Joseph Gonzalez, Quoc~V. Le, Chen Liang, Lluis{-}Miquel
  Munguia, Daniel Rothchild, David~R. So, Maud Texier, and Jeff Dean.
\newblock Carbon emissions and large neural network training.
\newblock {\em CoRR}, abs/2104.10350, 2021.

\bibitem{teacher-assistant}
Seyed{-}Iman Mirzadeh, Mehrdad Farajtabar, Ang Li, and Hassan Ghasemzadeh.
\newblock Improved knowledge distillation via teacher assistant: Bridging the
  gap between student and teacher.
\newblock {\em CoRR}, abs/1902.03393, 2019.

\bibitem{KDefficacy}
Jang~Hyun Cho and Bharath Hariharan.
\newblock On the efficacy of knowledge distillation.
\newblock In {\em Proceedings of the IEEE/CVF International Conference on
  Computer Vision}, pages 4794--4802, 2019.

\bibitem{rco}
Xiao Jin, Baoyun Peng, Yichao Wu, Yu~Liu, Jiaheng Liu, Ding Liang, Junjie Yan,
  and Xiaolin Hu.
\newblock Knowledge distillation via route constrained optimization.
\newblock In {\em Proceedings of the IEEE/CVF International Conference on
  Computer Vision}, pages 1345--1354, 2019.

\bibitem{follow-your-path}
Wenxian Shi, Yuxuan Song, Hao Zhou, Bohan Li, and Lei Li.
\newblock Follow your path: a progressive method for knowledge distillation.
\newblock In {\em Joint European Conference on Machine Learning and Knowledge
  Discovery in Databases}, pages 596--611. Springer, 2021.

\bibitem{skd}
Jia Guo, Minghao Chen, Yao Hu, Chen Zhu, Xiaofei He, and Deng Cai.
\newblock Reducing the teacher-student gap via spherical knowledge
  disitllation.
\newblock {\em arXiv preprint arXiv:2010.07485}, 2020.

\bibitem{Megatron-LM}
Mohammad Shoeybi, Mostofa Patwary, Raul Puri, Patrick LeGresley, Jared Casper,
  and Bryan Catanzaro.
\newblock Megatron-lm: Training multi-billion parameter language models using
  model parallelism.
\newblock {\em CoRR}, abs/1909.08053, 2019.

\bibitem{T5}
Colin Raffel, Noam Shazeer, Adam Roberts, Katherine Lee, Sharan Narang, Michael
  Matena, Yanqi Zhou, Wei Li, and Peter~J. Liu.
\newblock Exploring the limits of transfer learning with a unified text-to-text
  transformer.
\newblock {\em CoRR}, abs/1910.10683, 2019.

\bibitem{switch-transformer}
William Fedus, Barret Zoph, and Noam Shazeer.
\newblock Switch transformers: Scaling to trillion parameter models with simple
  and efficient sparsity.
\newblock {\em CoRR}, abs/2101.03961, 2021.

\bibitem{jurassic-1}
Jurassic-1: Technical details and evaluation.
\newblock
  \url{https://uploads-ssl.webflow.com/60fd4503684b466578c0d307/61138924626a6981ee09caf6_jurassic_tech_paper.pdf
  }.

\bibitem{gopher}
Scaling language models: Methods, analysis \& insights from training gopher.
\newblock
  \url{https://deepmind.com/research/publications/2021/scaling-language-models-methods-analysis-insights-from-training-gopher
  }.

\bibitem{megatron-nlg}
Using deepspeed and megatron to train megatron-turing nlg 530b, the world’s
  largest and most powerful generative language mode.
\newblock \url{https://developer.nvidia
  .com/blog/using-deepspeed-and-megatron-to-train-megatron-turing-nlg-530b
  -the-worlds-largest-and-most-powerful-generative-language-model/ }.

\bibitem{zeng2021pangu}
Wei Zeng, Xiaozhe Ren, Teng Su, Hui Wang, Yi~Liao, Zhiwei Wang, Xin Jiang,
  ZhenZhang Yang, Kaisheng Wang, Xiaoda Zhang, et~al.
\newblock Pangu-$alpha$: Large-scale autoregressive pretrained chinese language
  models with auto-parallel computation.
\newblock {\em arXiv preprint arXiv:2104.12369}, 2021.

\bibitem{wu2021yuan}
Shaohua Wu, Xudong Zhao, Tong Yu, Rongguo Zhang, Chong Shen, Hongli Liu, Feng
  Li, Hong Zhu, Jiangang Luo, Liang Xu, et~al.
\newblock Yuan 1.0: Large-scale pre-trained language model in zero-shot and
  few-shot learning.
\newblock {\em arXiv preprint arXiv:2110.04725}, 2021.

\bibitem{lewis2021base}
Mike Lewis, Shruti Bhosale, Tim Dettmers, Naman Goyal, and Luke Zettlemoyer.
\newblock Base layers: Simplifying training of large, sparse models.
\newblock {\em arXiv preprint arXiv:2103.16716}, 2021.

\bibitem{roller2021hash}
Stephen Roller, Sainbayar Sukhbaatar, Arthur Szlam, and Jason Weston.
\newblock Hash layers for large sparse models.
\newblock {\em arXiv preprint arXiv:2106.04426}, 2021.

\bibitem{zhang2021cpm}
Zhengyan Zhang, Yuxian Gu, Xu~Han, Shengqi Chen, Chaojun Xiao, Zhenbo Sun, Yuan
  Yao, Fanchao Qi, Jian Guan, Pei Ke, et~al.
\newblock Cpm-2: Large-scale cost-effective pre-trained language models.
\newblock {\em arXiv preprint arXiv:2106.10715}, 2021.

\bibitem{lin2021m6}
Junyang Lin, An~Yang, Jinze Bai, Chang Zhou, Le~Jiang, Xianyan Jia, Ang Wang,
  Jie Zhang, Yong Li, Wei Lin, et~al.
\newblock M6-10t: A sharing-delinking paradigm for efficient multi-trillion
  parameter pretraining.
\newblock {\em arXiv preprint arXiv:2110.03888}, 2021.

\bibitem{rise_risk_gpt3}
Matthew Hutson.
\newblock Robo-writers: the rise and risks of language-generating ai.
\newblock Website, 2021.
\newblock \url{https://www.nature.com/articles/d41586-021-00530-0}.

\bibitem{kingma2014adam}
Diederik~P Kingma and Jimmy Ba.
\newblock Adam: A method for stochastic optimization.
\newblock {\em arXiv preprint arXiv:1412.6980}, 2014.

\bibitem{huangGPipeEfficientTraining2019}
Yanping Huang, Youlong Cheng, Ankur Bapna, Orhan Firat, Mia~Xu Chen, Dehao
  Chen, HyoukJoong Lee, Jiquan Ngiam, Quoc~V. Le, Yonghui Wu, and Zhifeng Chen.
\newblock {{GPipe}}: Efficient {{Training}} of {{Giant Neural Networks}} using
  {{Pipeline Parallelism}}.

\bibitem{liTeraPipeTokenLevelPipeline2021}
Zhuohan Li, Siyuan Zhuang, Shiyuan Guo, Danyang Zhuo, Hao Zhang, Dawn Song, and
  Ion Stoica.
\newblock {{TeraPipe}}: Token-{{Level Pipeline Parallelism}} for {{Training
  Large}}-{{Scale Language Models}}.

\bibitem{narayanan2021efficient}
Deepak Narayanan, Mohammad Shoeybi, Jared Casper, Patrick LeGresley, Mostofa
  Patwary, Vijay~Anand Korthikanti, Dmitri Vainbrand, Prethvi Kashinkunti,
  Julie Bernauer, Bryan Catanzaro, et~al.
\newblock Efficient large-scale language model training on gpu clusters.
\newblock {\em arXiv preprint arXiv:2104.04473}, 2021.

\bibitem{yao2019plan}
Lili Yao, Nanyun Peng, Ralph Weischedel, Kevin Knight, Dongyan Zhao, and Rui
  Yan.
\newblock Plan-and-write: Towards better automatic storytelling.
\newblock In {\em Proceedings of the AAAI Conference on Artificial
  Intelligence}, volume~33, pages 7378--7385, 2019.

\bibitem{wu2020controllable}
Zeqiu Wu, Michel Galley, Chris Brockett, Yizhe Zhang, Xiang Gao, Chris Quirk,
  Rik Koncel-Kedziorski, Jianfeng Gao, Hannaneh Hajishirzi, Mari Ostendorf,
  et~al.
\newblock A controllable model of grounded response generation.
\newblock {\em arXiv preprint arXiv:2005.00613}, 2020.

\bibitem{kreps2020all}
Sarah Kreps, R~Miles McCain, and Miles Brundage.
\newblock All the news that’s fit to fabricate: Ai-generated text as a tool
  of media misinformation.
\newblock {\em Journal of Experimental Political Science}, pages 1--14, 2020.

\bibitem{radford2019language}
Alec Radford, Jeffrey Wu, Rewon Child, David Luan, Dario Amodei, and Ilya
  Sutskever.
\newblock Language models are unsupervised multitask learners.
\newblock {\em OpenAI Blog}, 1(8), 2019.

\bibitem{zellers2019defending}
Rowan Zellers, Ari Holtzman, Hannah Rashkin, Yonatan Bisk, Ali Farhadi,
  Franziska Roesner, and Yejin Choi.
\newblock Defending against neural fake news.
\newblock {\em arXiv preprint arXiv:1905.12616}, 2019.

\bibitem{NitishShirishKeskar2019CTRLAC}
Nitish~Shirish Keskar, Bryan McCann, Lav~R. Varshney, Caiming Xiong, and
  Richard Socher.
\newblock Ctrl: A conditional transformer language model for controllable
  generation.
\newblock {\em arXiv: Computation and Language}, 2019.

\bibitem{BenjaminSchiller2020AspectControlledNA}
Benjamin Schiller, Johannes Daxenberger, and Iryna Gurevych.
\newblock Aspect-controlled neural argument generation.
\newblock {\em arXiv: Computation and Language}, 2020.

\bibitem{AlexisRoss2021TailorGA}
Alexis Ross, Tongshuang Wu, Hao Peng, Matthew~E. Peters, and Matt Gardner.
\newblock Tailor: Generating and perturbing text with semantic controls.
\newblock {\em arXiv: Computation and Language}, 2021.

\bibitem{SumanthDathathri2019PlugAP}
Sumanth Dathathri, Andrea Madotto, Janice Lan, Jane Hung, Eric Frank, Piero
  Molino, Jason Yosinski, and Rosanne Liu.
\newblock Plug and play language models: A simple approach to controlled text
  generation.
\newblock {\em arXiv: Computation and Language}, 2019.

\bibitem{AlvinChan2020CoConAS}
Alvin Chan, Yew-Soon Ong, Bill Pung, Aston Zhang, and Jie Fu.
\newblock Cocon: A self-supervised approach for controlled text generation.
\newblock {\em arXiv: Computation and Language}, 2020.

\bibitem{pkd}
Siqi Sun, Yu~Cheng, Zhe Gan, and Jingjing Liu.
\newblock Patient knowledge distillation for {BERT} model compression.
\newblock In {\em Proceedings of the 2019 Conference on Empirical Methods in
  Natural Language Processing and the 9th International Joint Conference on
  Natural Language Processing (EMNLP-IJCNLP)}, pages 4323--4332, Hong Kong,
  China, November 2019. Association for Computational Linguistics.

\bibitem{mobilebert}
Zhiqing Sun, Hongkun Yu, Xiaodan Song, Renjie Liu, Yiming Yang, and Denny Zhou.
\newblock {M}obile{BERT}: a compact task-agnostic {BERT} for resource-limited
  devices.
\newblock In {\em Proceedings of the 58th Annual Meeting of the Association for
  Computational Linguistics}, pages 2158--2170, Online, July 2020. Association
  for Computational Linguistics.

\bibitem{well-read}
Iulia Turc, Ming-Wei Chang, Kenton Lee, and Kristina Toutanova.
\newblock Well-read students learn better: On the importance of pre-training
  compact models.
\newblock {\em arXiv preprint arXiv:1908.08962}, 2019.

\bibitem{distill-bert}
Xingkai Ren, Ronghua Shi, and Fangfang Li.
\newblock Distill bert to traditional models in chinese machine reading
  comprehension (student abstract).
\newblock In {\em Proceedings of the AAAI Conference on Artificial
  Intelligence}, volume~34, pages 13901--13902, 2020.

\bibitem{sixteen}
Paul Michel, Omer Levy, and Graham Neubig.
\newblock Are sixteen heads really better than one?
\newblock In Hanna~M. Wallach, Hugo Larochelle, Alina Beygelzimer, Florence
  d'Alch{\'{e}}{-}Buc, Emily~B. Fox, and Roman Garnett, editors, {\em Advances
  in Neural Information Processing Systems 32: Annual Conference on Neural
  Information Processing Systems 2019, NeurIPS 2019, December 8-14, 2019,
  Vancouver, BC, Canada}, pages 14014--14024, 2019.

\bibitem{akd}
Aref Jafari, Mehdi Rezagholizadeh, Pranav Sharma, and Ali Ghodsi.
\newblock Annealing knowledge distillation.
\newblock {\em arXiv preprint arXiv:2104.07163}, 2021.

\bibitem{joshi2020spanbert}
Mandar Joshi, Danqi Chen, Yinhan Liu, Daniel~S Weld, Luke Zettlemoyer, and Omer
  Levy.
\newblock Spanbert: Improving pre-training by representing and predicting
  spans.
\newblock {\em Transactions of the Association for Computational Linguistics},
  8:64--77, 2020.

\bibitem{sun2020ernie}
Yu~Sun, Shuohuan Wang, Yukun Li, Shikun Feng, Hao Tian, Hua Wu, and Haifeng
  Wang.
\newblock Ernie 2.0: A continual pre-training framework for language
  understanding.
\newblock In {\em Proceedings of the AAAI Conference on Artificial
  Intelligence}, volume~34, pages 8968--8975, 2020.

\bibitem{dai2019transformer}
Zihang Dai, Zhilin Yang, Yiming Yang, William~W Cohen, Jaime Carbonell, Quoc~V
  Le, and Ruslan Salakhutdinov.
\newblock Transformer-xl: Attentive language models beyond a fixed-length
  context.
\newblock {\em arXiv preprint arXiv:1901.02860}, 2019.

\bibitem{yang2019xlnet}
Zhilin Yang, Zihang Dai, Yiming Yang, Jaime Carbonell, Ruslan Salakhutdinov,
  and Quoc~V Le.
\newblock Xlnet: Generalized autoregressive pretraining for language
  understanding.
\newblock {\em arXiv preprint arXiv:1906.08237}, 2019.

\bibitem{DBLP:journals/corr/abs-2004-14996}
He~Bai, Peng Shi, Jimmy Lin, Luchen Tan, Kun Xiong, Wen Gao, and Ming Li.
\newblock Segabert: Pre-training of segment-aware {BERT} for language
  understanding.
\newblock {\em CoRR}, abs/2004.14996, 2020.

\bibitem{ding2020ernie}
Siyu Ding, Junyuan Shang, Shuohuan Wang, Yu~Sun, Hao Tian, Hua Wu, and Haifeng
  Wang.
\newblock Ernie-doc: The retrospective long-document modeling transformer.
\newblock {\em arXiv preprint arXiv:2012.15688}, 2020.

\bibitem{megatron}
Mohammad Shoeybi, Mostofa Patwary, Raul Puri, Patrick LeGresley, Jared Casper,
  and Bryan Catanzaro.
\newblock Megatron-lm: Training multi-billion parameter language models using
  model parallelism.
\newblock {\em arXiv preprint arXiv:1909.08053}, 2019.

\bibitem{rajbhandariZeROMemoryOptimizations2020}
Samyam Rajbhandari, Jeff Rasley, Olatunji Ruwase, and Yuxiong He.
\newblock {{ZeRO}}: Memory {{Optimizations Toward Training Trillion Parameter
  Models}}.

\bibitem{DBLP:journals/corr/abs-2012-15688}
Siyu Ding, Junyuan Shang, Shuohuan Wang, Yu~Sun, Hao Tian, Hua Wu, and Haifeng
  Wang.
\newblock {ERNIE-DOC:} the retrospective long-document modeling transformer.
\newblock {\em CoRR}, abs/2012.15688, 2020.

\bibitem{horovod}
Alexander Sergeev and Mike Del~Balso.
\newblock Horovod: fast and easy distributed deep learning in tensorflow.
\newblock {\em arXiv preprint arXiv:1802.05799}, 2018.

\bibitem{liPyTorchDistributedExperiences2020a}
Shen Li, Yanli Zhao, Rohan Varma, Omkar Salpekar, Pieter Noordhuis, Teng Li,
  Adam Paszke, Jeff Smith, Brian Vaughan, Pritam Damania, and Soumith Chintala.
\newblock {{PyTorch}} distributed: Experiences on accelerating data parallel
  training.
\newblock 13(12):3005--3018.

\bibitem{tensorflowmesh}
Noam Shazeer, Youlong Cheng, Niki Parmar, Dustin Tran, Ashish Vaswani, Penporn
  Koanantakool, Peter Hawkins, HyoukJoong Lee, Mingsheng Hong, Cliff Young,
  et~al.
\newblock Mesh-tensorflow: Deep learning for supercomputers.
\newblock {\em arXiv preprint arXiv:1811.02084}, 2018.

\bibitem{pipedream}
Aaron Harlap, Deepak Narayanan, Amar Phanishayee, Vivek Seshadri, Nikhil
  Devanur, Greg Ganger, and Phil Gibbons.
\newblock Pipedream: Fast and efficient pipeline parallel dnn training.
\newblock {\em arXiv preprint arXiv:1806.03377}, 2018.

\bibitem{pipedream2bw}
Deepak Narayanan, Aaron Harlap, Amar Phanishayee, Vivek Seshadri, Nikhil~R.
  Devanur, Gregory~R. Ganger, Phillip~B. Gibbons, and Matei Zaharia.
\newblock Pipedream: Generalized pipeline parallelism for dnn training.
\newblock SOSP '19, page 1–15, New York, NY, USA, 2019. Association for
  Computing Machinery.

\bibitem{narayananMemoryEfficientPipelineParallelDNN2021}
Deepak Narayanan, Amar Phanishayee, Kaiyu Shi, Xie Chen, and Matei Zaharia.
\newblock Memory-{{Efficient Pipeline}}-{{Parallel DNN Training}}.

\bibitem{ao2021endtoend}
Yulong Ao, Zhihua Wu, Dianhai Yu, Weibao Gong, Zhiqing Kui, Minxu Zhang,
  Zilingfeng Ye, Liang Shen, Yanjun Ma, Tian Wu, Haifeng Wang, Wei Zeng, and
  Chao Yang.
\newblock End-to-end adaptive distributed training on paddlepaddle, 2021.

\bibitem{vaswani2017attention}
Ashish Vaswani, Noam Shazeer, Niki Parmar, Jakob Uszkoreit, Llion Jones,
  Aidan~N Gomez, {\L}ukasz Kaiser, and Illia Polosukhin.
\newblock Attention is all you need.
\newblock In {\em Advances in neural information processing systems}, pages
  5998--6008, 2017.

\bibitem{xu2021fewclue}
Liang Xu, Xiaojing Lu, Chenyang Yuan, Xuanwei Zhang, Huilin Xu, Hu~Yuan, Guoao
  Wei, Xiang Pan, Xin Tian, Libo Qin, and Hu~Hai.
\newblock Fewclue: A chinese few-shot learning evaluation benchmark, 2021.

\bibitem{li2018cote}
Yanzeng Li, Tingwen Liu, Diying Li, Quangang Li, Jinqiao Shi, and Yanqiu Wang.
\newblock Character-based bilstm-crf incorporating pos and dictionaries for
  chinese opinion target extraction.
\newblock In {\em Asian Conference on Machine Learning}, pages 518--533. PMLR,
  2018.

\bibitem{xu2020clue}
Liang Xu, Hai Hu, Xuanwei Zhang, Lu~Li, Chenjie Cao, Yudong Li, Yechen Xu, Kai
  Sun, Dian Yu, Cong Yu, et~al.
\newblock Clue: A chinese language understanding evaluation benchmark.
\newblock {\em arXiv preprint arXiv:2004.05986}, 2020.

\bibitem{li2019finre}
Ziran Li, Ning Ding, Zhiyuan Liu, Haitao Zheng, and Ying Shen.
\newblock Chinese relation extraction with multi-grained information and
  external linguistic knowledge.
\newblock In {\em Proceedings of the 57th Annual Meeting of the Association for
  Computational Linguistics}, pages 4377--4386, 2019.

\bibitem{xu2017discourse}
Jingjing Xu, Ji~Wen, Xu~Sun, and Qi~Su.
\newblock A discourse-level named entity recognition and relation extraction
  dataset for chinese literature text.
\newblock {\em arXiv preprint arXiv:1711.07010}, 2017.

\bibitem{liu2018lcqmc}
Xin Liu, Qingcai Chen, Chong Deng, Huajun Zeng, Jing Chen, Dongfang Li, and
  Buzhou Tang.
\newblock Lcqmc: A large-scale chinese question matching corpus.
\newblock In {\em Proceedings of the 27th International Conference on
  Computational Linguistics}, pages 1952--1962, 2018.

\bibitem{yang2019paws}
Yinfei Yang, Yuan Zhang, Chris Tar, and Jason Baldridge.
\newblock Paws-x: A cross-lingual adversarial dataset for paraphrase
  identification.
\newblock {\em arXiv preprint arXiv:1908.11828}, 2019.

\bibitem{chen2018bq}
Jing Chen, Qingcai Chen, Xin Liu, Haijun Yang, Daohe Lu, and Buzhou Tang.
\newblock The bq corpus: A large-scale domain-specific chinese corpus for
  sentence semantic equivalence identification.
\newblock In {\em Proceedings of the 2018 Conference on Empirical Methods in
  Natural Language Processing}, pages 4946--4951, 2018.

\bibitem{co2019iflytek}
LTD~IFLYTEK CO.
\newblock Iflytek: a multiple categories chinese text classifier. competition
  official website, 2019.

\bibitem{liu2018matching}
Bang Liu, Di~Niu, Haojie Wei, Jinghong Lin, Yancheng He, Kunfeng Lai, and
  Yu~Xu.
\newblock Matching article pairs with graphical decomposition and convolutions.
\newblock {\em arXiv preprint arXiv:1802.07459}, 2018.

\bibitem{zhang2017cmedqa}
Sheng Zhang, Xin Zhang, Hui Wang, Jiajun Cheng, Pei Li, and Zhaoyun Ding.
\newblock Chinese medical question answer matching using end-to-end
  character-level multi-scale cnns.
\newblock {\em Applied Sciences}, 7(8):767, 2017.

\bibitem{zhang2018multi}
Sheng Zhang, Xin Zhang, Hui Wang, Lixiang Guo, and Shanshan Liu.
\newblock Multi-scale attentive interaction networks for chinese medical
  question answer selection.
\newblock {\em IEEE Access}, 6:74061--74071, 2018.

\bibitem{li2016webqa}
Peng Li, Wei Li, Zhengyan He, Xuguang Wang, Ying Cao, Jie Zhou, and Wei Xu.
\newblock Dataset and neural recurrent sequence labeling model for open-domain
  factoid question answering.
\newblock {\em arXiv preprint arXiv:1607.06275}, 2016.

\bibitem{cui2016pdcft}
Yiming Cui, Ting Liu, Zhipeng Chen, Shijin Wang, and Guoping Hu.
\newblock Consensus attention-based neural networks for chinese reading
  comprehension.
\newblock {\em arXiv preprint arXiv:1607.02250}, 2016.

\bibitem{cui2017dataset}
Yiming Cui, Ting Liu, Zhipeng Chen, Wentao Ma, Shijin Wang, and Guoping Hu.
\newblock Dataset for the first evaluation on chinese machine reading
  comprehension.
\newblock {\em arXiv preprint arXiv:1709.08299}, 2017.

\bibitem{cui2020sentence}
Yiming Cui, Ting Liu, Ziqing Yang, Zhipeng Chen, Wentao Ma, Wanxiang Che,
  Shijin Wang, and Guoping Hu.
\newblock A sentence cloze dataset for chinese machine reading comprehension.
\newblock {\em arXiv preprint arXiv:2004.03116}, 2020.

\bibitem{zheng2019chid}
Chujie Zheng, Minlie Huang, and Aixin Sun.
\newblock Chid: A large-scale chinese idiom dataset for cloze test.
\newblock {\em arXiv preprint arXiv:1906.01265}, 2019.

\bibitem{ge2021chinese}
Huibin Ge, Chenxi Sun, Deyi Xiong, and Qun Liu.
\newblock Chinese wplc: A chinese dataset for evaluating pretrained language
  models on word prediction given long-range context.
\newblock In {\em Proceedings of the 2021 Conference on Empirical Methods in
  Natural Language Processing}, pages 3770--3778, 2021.

\bibitem{shao2018drcd}
Chih~Chieh Shao, Trois Liu, Yuting Lai, Yiying Tseng, and Sam Tsai.
\newblock Drcd: a chinese machine reading comprehension dataset.
\newblock {\em arXiv preprint arXiv:1806.00920}, 2018.

\bibitem{he2017dureader}
Wei He, Kai Liu, Jing Liu, Yajuan Lyu, Shiqi Zhao, Xinyan Xiao, Yuan Liu,
  Yizhong Wang, Hua Wu, Qiaoqiao She, et~al.
\newblock Dureader: a chinese machine reading comprehension dataset from
  real-world applications.
\newblock {\em arXiv preprint arXiv:1711.05073}, 2017.

\bibitem{tang2020dureaderrobust}
Hongxuan Tang, Jing Liu, Hongyu Li, Yu~Hong, Hua Wu, and Haifeng Wang.
\newblock Dureaderrobust: A chinese dataset towards evaluating the robustness
  of machine reading comprehension models.
\newblock {\em arXiv preprint arXiv:2004.11142}, 2020.

\bibitem{sun2020c3}
Kai Sun, Dian Yu, Dong Yu, and Claire Cardie.
\newblock Investigating prior knowledge for challenging chinese machine reading
  comprehension.
\newblock {\em Transactions of the Association for Computational Linguistics},
  8:141--155, 2020.

\bibitem{DBLP:journals/corr/abs-1810-07366}
Yiming Cui, Ting Liu, Li~Xiao, Zhipeng Chen, Wentao Ma, Wanxiang Che, Shijin
  Wang, and Guoping Hu.
\newblock A span-extraction dataset for chinese machine reading comprehension.
\newblock {\em CoRR}, abs/1810.07366, 2018.

\bibitem{xiao2018cail2018}
Chaojun Xiao, Haoxi Zhong, Zhipeng Guo, Cunchao Tu, Zhiyuan Liu, Maosong Sun,
  Yansong Feng, Xianpei Han, Zhen Hu, Heng Wang, et~al.
\newblock Cail2018: A large-scale legal dataset for judgment prediction.
\newblock {\em arXiv preprint arXiv:1807.02478}, 2018.

\bibitem{xu-etal-2021-blow}
Canwen Xu, Wangchunshu Zhou, Tao Ge, Ke~Xu, Julian McAuley, and Furu Wei.
\newblock Blow the dog whistle: A {C}hinese dataset for cant understanding with
  common sense and world knowledge.
\newblock In {\em Proceedings of the 2021 Conference of the North American
  Chapter of the Association for Computational Linguistics: Human Language
  Technologies}, pages 2139--2145, Online, June 2021. Association for
  Computational Linguistics.

\bibitem{tian2020skep}
Hao Tian, Can Gao, Xinyan Xiao, Hao Liu, Bolei He, Hua Wu, Haifeng Wang, and
  Feng Wu.
\newblock Skep: Sentiment knowledge enhanced pre-training for sentiment
  analysis.
\newblock {\em arXiv preprint arXiv:2005.05635}, 2020.

\bibitem{cui2019pre}
Yiming Cui, Wanxiang Che, Ting Liu, Bing Qin, Ziqing Yang, Shijin Wang, and
  Guoping Hu.
\newblock Pre-training with whole word masking for chinese bert.
\newblock {\em arXiv preprint arXiv:1906.08101}, 2019.

\bibitem{lan2019albert}
Zhenzhong Lan, Mingda Chen, Sebastian Goodman, Kevin Gimpel, Piyush Sharma, and
  Radu Soricut.
\newblock Albert: A lite bert for self-supervised learning of language
  representations.
\newblock {\em arXiv preprint arXiv:1909.11942}, 2019.

\bibitem{cui2020revisiting}
Yiming Cui, Wanxiang Che, Ting Liu, Bing Qin, Shijin Wang, and Guoping Hu.
\newblock Revisiting pre-trained models for chinese natural language
  processing.
\newblock {\em arXiv preprint arXiv:2004.13922}, 2020.

\bibitem{song2021zen}
Yan Song, Tong Zhang, Yonggang Wang, and Kai-Fu Lee.
\newblock Zen 2.0: Continue training and adaption for n-gram enhanced text
  encoders.
\newblock {\em arXiv preprint arXiv:2105.01279}, 2021.

\bibitem{cui2020chinese}
Xiongtao Cui and Jungang Han.
\newblock Chinese medical question answer matching based on interactive
  sentence representation learning.
\newblock {\em arXiv preprint arXiv:2011.13573}, 2020.

\bibitem{schick2020exploiting}
Timo Schick and Hinrich Sch{\"u}tze.
\newblock Exploiting cloze questions for few shot text classification and
  natural language inference.
\newblock {\em arXiv preprint arXiv:2001.07676}, 2020.

\bibitem{sun2021nsp}
Yi~Sun, Yu~Zheng, Chao Hao, and Hangping Qiu.
\newblock Nsp-bert: A prompt-based zero-shot learner through an original
  pre-training task--next sentence prediction.
\newblock {\em arXiv preprint arXiv:2109.03564}, 2021.

\bibitem{AriHoltzman2021SurfaceFC}
Ari Holtzman, Peter West, Vered Shwartz, Yejin Choi, and Luke Zettlemoyer.
\newblock Surface form competition: Why the highest probability answer isn't
  always right.
\newblock {\em arXiv: Computation and Language}, 2021.

\bibitem{kuhn1955hungarian}
Harold~W Kuhn.
\newblock The hungarian method for the assignment problem.
\newblock {\em Naval research logistics quarterly}, 2(1-2):83--97, 1955.

\bibitem{conneau2018xnli}
Alexis Conneau, Guillaume Lample, Ruty Rinott, Adina Williams, Samuel~R Bowman,
  Holger Schwenk, and Veselin Stoyanov.
\newblock Xnli: Evaluating cross-lingual sentence representations.
\newblock {\em arXiv preprint arXiv:1809.05053}, 2018.

\bibitem{song6chnsenticorp}
TAN Song-bo.
\newblock Chnsenticorp.

\bibitem{cmrc2018}
Yiming Cui, Ting Liu, Wanxiang Che, Li~Xiao, Zhipeng Chen, Wentao Ma, Shijin
  Wang, and Guoping Hu.
\newblock A span-extraction dataset for chinese machine reading comprehension.
\newblock {\em arXiv preprint arXiv:1810.07366}, 2018.

\bibitem{liu2019roberta}
Yinhan Liu, Myle Ott, Naman Goyal, Jingfei Du, Mandar Joshi, Danqi Chen, Omer
  Levy, Mike Lewis, Luke Zettlemoyer, and Veselin Stoyanov.
\newblock Roberta: A robustly optimized bert pretraining approach, 2019.
\newblock cite arxiv:1907.11692.

\end{thebibliography}

\newpage

\end{document}